\documentclass[10pt,journal,compsoc]{IEEEtran}
\usepackage{hyperref}
\hypersetup{hypertex=true,colorlinks=true,linkcolor=green}
\pdfoutput=1

\usepackage{ragged2e}
\usepackage{url}
\usepackage{setspace}
\usepackage{xspace}
\usepackage{tablefootnote}
\usepackage{multirow}
\usepackage{booktabs}
\usepackage{color}
\usepackage{graphicx}
\usepackage{bm}
\usepackage{tikz}
\usepackage{comment}
\usepackage{amsmath,amssymb,amsfonts}
\usepackage{color}
\usepackage{multirow}
\usepackage{setspace}
\usepackage{booktabs}

\usepackage{algorithm}
\usepackage{colortbl}
\usepackage{algorithmic}
\usepackage{subfigure}
\usepackage{makecell}
\newtheorem{remark}{Remark}
\usepackage{epstopdf}

\ifCLASSOPTIONcompsoc
  \usepackage[nocompress]{cite}
\else
  \usepackage{cite}
\fi
\ifCLASSINFOpdf
\else
\fi
\begin{document}
%
\title{Learning Task-preferred Inference Routes for Gradient De-conflict in Multi-output DNNs}
%
%
%
%

\author{Yi Sun, Xiaochang Hu, Xin Xu*,~\IEEEmembership{IEEE Senior Member}, Jian Li*, Yifei Shi, and Ling-Li Zeng
\IEEEcompsocitemizethanks{
\IEEEcompsocthanksitem Yi Sun is with Chinese Academy of Sciences, Beijing 100190; Xiaochang Hu is with Beijing Institute of Basic Medical Sciences, Beijing 100850, China; Xin Xu, Jian Li, Yifei Shi and Ling-Li Zeng are with the College of Intelligence Science and Technology, National University of Defense Technology, China,410000.\protect\\
Corresponding author: Xin Xu and Jian Li. E-mail: xinxu@nudt.edu.cn, lijian@nudt.edu.cn}}

%
%

\markboth{\fontsize{5.0pt}{\baselineskip}\selectfont This work has been submitted to the IEEE for possible publication. Copyright may be transferred without notice, after which this version may no longer be accessible.}%
{Shell \MakeLowercase{\textit{et al.}}: Bare Demo of IEEEtran.cls for Computer Society Journals}

\IEEEtitleabstractindextext{%
\begin{abstract}
\justifying
Multi-output deep neural networks (MONs) contain multiple output branches of various tasks, and these tasks typically share partial network filters, resulting in entangled inference routes between different tasks within the networks. Due to the divergent optimization objectives, the task gradients during training usually interfere with each other along the shared routes, which decreases the overall model performance. To address this issue, we propose a novel gradient de-conflict algorithm named DR-MGF (Dynamic Routes and Meta-weighted Gradient Fusion). Different from existing de-conflict methods, DR-MGF achieves gradient de-conflict in MONs by learning task-preferred inference routes. The proposed method is motivated by our experimental findings that the shared filters are not equally important for different tasks. By designing  learnable task-specific importance variables, DR-MGF evaluates the importance of filters for different tasks. Through making the dominance of tasks over filters proportional to the task-specific importance of filters, DR-MGF can effectively reduce inter-task interference. These task-specific importance variables ultimately determine task-preferred inference routes at the end of training iterations. Extensive experimental results on CIFAR, ImageNet, and NYUv2 demonstrate that DR-MGF outperforms existing de-conflict methods. Furthermore, DR-MGF can be extended to general MONs without modifying the overall network structures.
\end{abstract}

}

\maketitle

\IEEEdisplaynontitleabstractindextext

%
\IEEEpeerreviewmaketitle

\IEEEraisesectionheading{\section{Introduction}\label{sec:introduction}}

\begin{figure}[h]
\centering
\subfigure[]{\includegraphics[height=37mm]{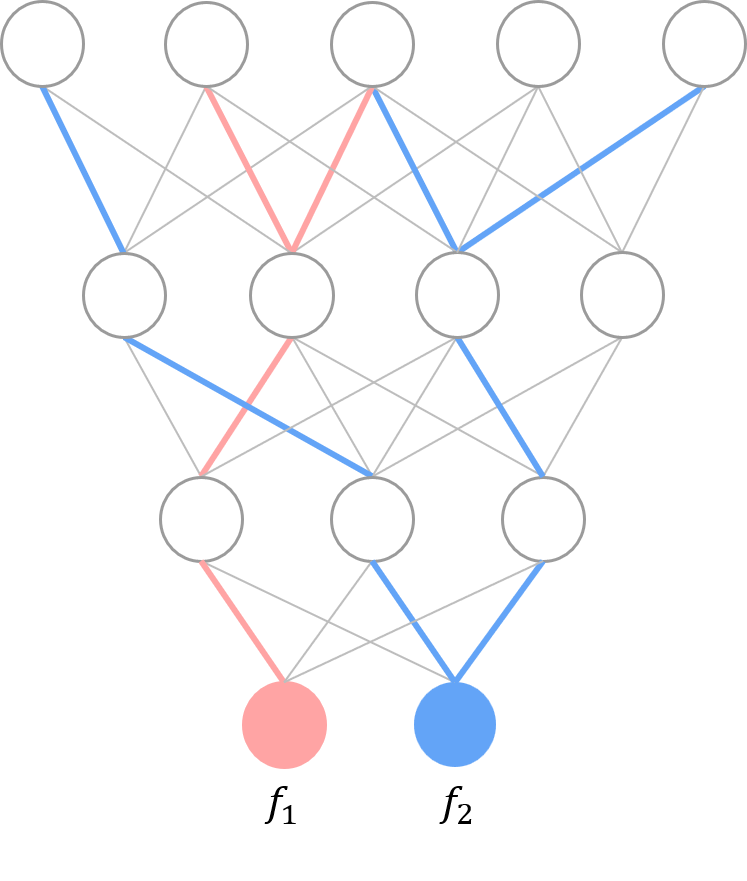}}
\subfigure[]{\includegraphics[height=40mm]{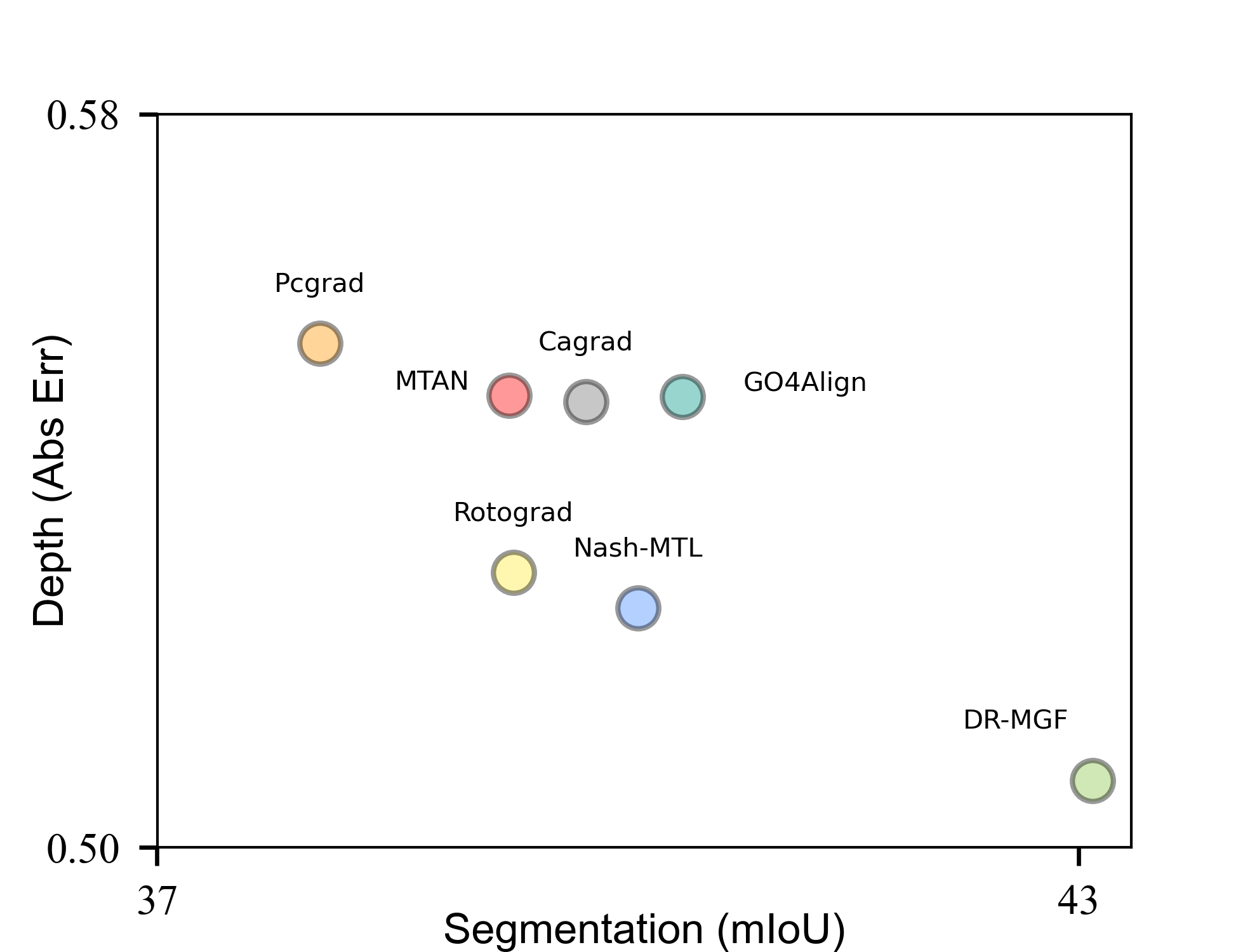}}\\
\subfigure[]{\includegraphics[width=0.45\textwidth]{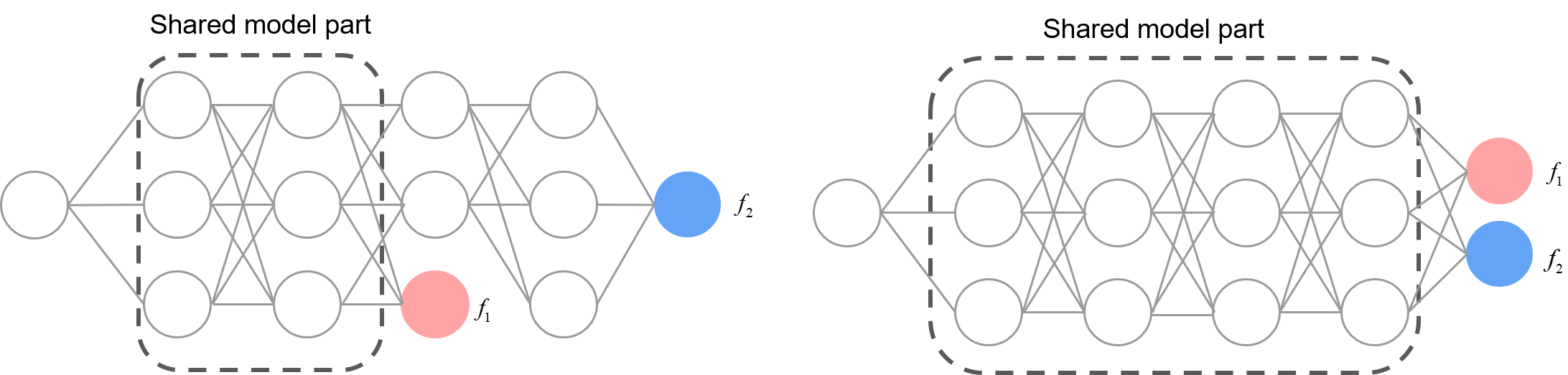}}
\caption{The proposed DR-MGF can effectively tackle the Multi-Objective Optimization (MOO)\cite{gunantara2018review} problem of MONs by adaptively identifying task-preferred inference routes. (a) Identifying task-preferred inference routes for different tasks, where $f_{1}$ and $f_{2}$ denote objective functions; (b) Multi-task prediction performance of Segnet-MTAN\cite{liu2019end} on NYUv2\cite{couprie2013indoor}, when using different optimization methods (Bottom-right is better); (c) Two types of multi-output networks. Multi-task networks (left) are designed for multiple task prediction, where the output branches are responsible for different tasks. Multi-exit networks (right) are applied to achieve depth-adaptive inference\cite{han2021dynamic}, where the output branches typically perform the same task but are positioned at different depths within the network architecture.}
\label{teaser1}
\end{figure}

\IEEEPARstart{O}{ver} the past decades, Deep Neural Networks (DNNs) have become fundamental technologies across various research domains such as autonomous driving\cite{chen2022milestones}, natural language processing\cite{liu2023pre}, protein design\cite{notin2024machine}. To fully leverage the capacity of DNNs, numerous multi-task networks\cite{vandenhende2021multi,huang2022curriculum,ding2020multi,guo2020learning,sun2020adashare,badrinarayanan2017segnet} and multi-exit networks\cite{huang2017multi,kaya2019shallow,passalis2020efficient,jie2019anytime} have been proposed. Compared with single-output models, multi-task networks are capable of performing multiple predictions in one inference stage, and multi-exit networks can dynamically adjust output depth according to the complexity of inputs at test time (see Fig. \ref{teaser1}(c)). Since both types of networks have a series of output branches with multiple learning objectives, this work collectively refers to multi-task and multi-exit networks as Multi-Output Networks (MONs) for unified description. Unless otherwise specified, we will refer to the exits of multi-exit networks as tasks in the following descriptions.\par

\begin{figure*}[t]
\centering
\includegraphics[width=1\textwidth]{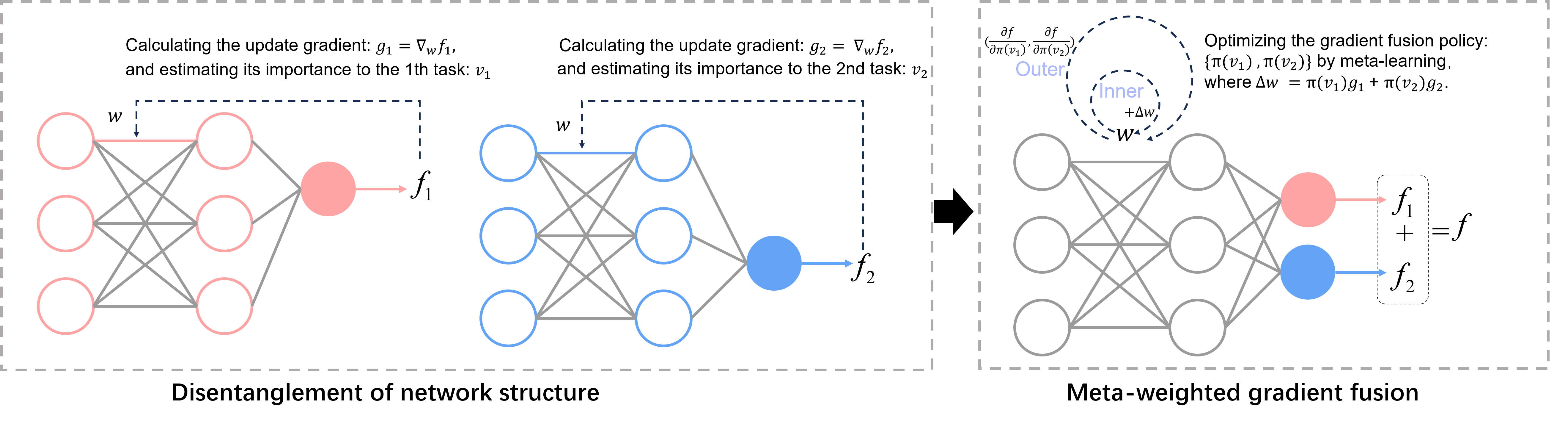}
\caption{Overview of DR-MGF. DR-MGF follows a disentanglement-and-fusion paradigm in each training epoch. In the disentanglement stage, DR-MGF calculates gradients of each task independently and learns task-specific importance variables of filters. In the fusion stage, DR-MGF employs a meta-weighted gradient fusion policy to fuse the gradients, enabling tasks to dominate the optimization of their preferred filters.}
\label{fig1}
\end{figure*}

The training process of MONs is a Multi-Objective Optimization (MOO) problem, which aims to find the optimal solution for multiple learning objectives\cite{shen2024go4align,gunantara2018review,martinez2020minimax}. However, different output branches in MONs usually share a large number of filters. When tasks interfere with each other during training, those shared filters will receive conflicting gradients from different tasks\cite{yu2020gradient} (see Fig. \ref{conflictproblem}(a)), degrading the overall performance of MONs. The inter-task interference typically arises from two forms of diversity among task gradients: differences in gradient magnitudes and inconsistencies in gradient directions. The former typically causes tasks with larger gradients to dominate the optimization process, consequently disrupting the optimization of tasks with smaller gradients. The latter can result in gradients of different tasks cancelling each other out, thus hindering the convergence for some or all tasks \cite{javaloy2021rotograd}. \par

Recent efforts to alleviate inter-task interference can be categorized into two categories: \textit{magnitude-adjustment de-conflict methods} and \textit{direction-projection de-conflict methods}. The former adjusts the gradient magnitude of tasks by weighting the task loss functions\cite{liuauto} or directly manipulating the gradient norm\cite{chen2018gradnorm,liu2021towards,yang2022adatask}. Differently, \textit{direction-projection} methods\cite{yu2020gradient,javaloy2021rotograd,liu2021conflict,navon2022multi} address the conflict by adjusting the directions of joint gradients during training. For example, numerous scalarization-based MOO methods\cite{shen2024go4align, gunantara2018review} such as Cagrad\cite{liu2021conflict} and Nash-MTL\cite{navon2022multi} have been proposed to find solutions that balance multiple objectives. \par
Despite existing progress, prior approaches\cite{liu2021conflict,navon2022multi,martinez2020minimax} typically treat the networks as a single unit, and thus ignore the varying importance of shared filters across tasks. This usually leads to a performance trade-off, where certain tasks are improved at the expense of others. Our experiments in this work reveal that the shared filters are not always equally important for different tasks, indicating that inter-task interference can be addressed by learning task-preferred inference routes (Fig. \ref{teaser1}). In other words, the gradient conflict problem of MONs can be tackled by enabling tasks to dominate the optimization of filters along their preferred inference routes. To the best of our knowledge, no existing work has explored gradient de-conflict solutions from this perspective, and doesn't require to modify the overall network architecture.\par

Therefore, we propose a novel gradient de-conflict method named DR-MGF (Dynamic Routes and Meta-weighted Gradient Fusion) for training MONs, which can effectively reduce the inter-task interference by aligning the dominance of tasks over filters with the task-specific importance of filters. In each training epoch, DR-MGF employs a two-stage learning paradigm comprising disentanglement and fusion as shown in Fig. \ref{fig1}. In the disentanglement stage, DR-MGF calculates gradients for each task independently and learns task-specific importance variables of filters\footnote{The task-specific importance variables represent the importance of filters for each task}. In the fusion stage, DR-MGF employs a meta-weighted gradient fusion algorithm, which is proposed in our previous works\cite{sun2022meta}, to synthesize the gradients of all tasks in a bi-level optimization manner. By following this two-stage process, DR-MGF enables tasks to dominate the optimization of their preferred filters, and alleviates inter-task interference. Finally, the learned task-specific importance variables determine task-preferred inference routes without modifying the overall network structures.\par
\begin{figure*}[t]
\centering
\subfigure[]{\includegraphics[trim=0 2 0 5,clip,height=60mm]{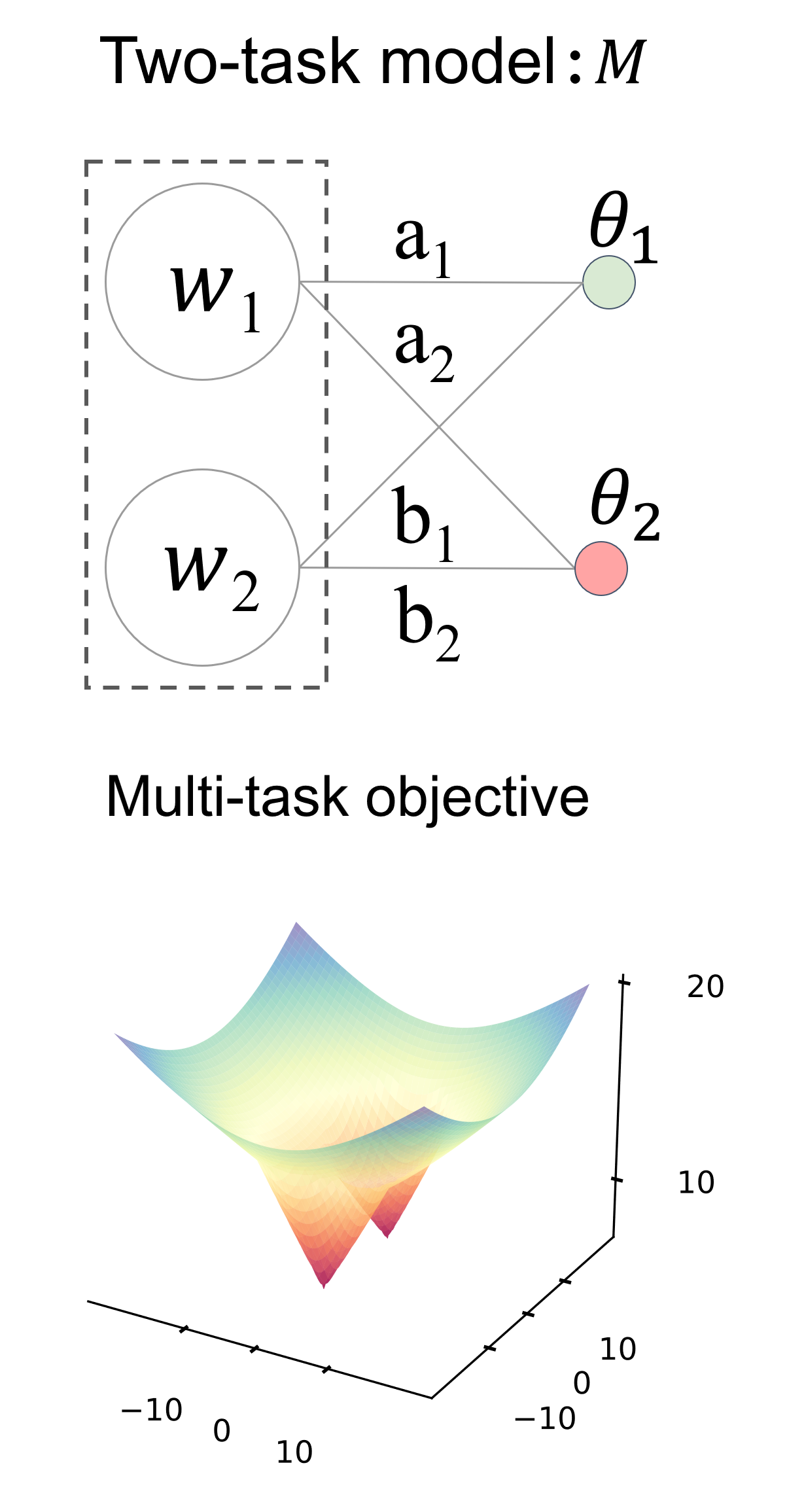}}
\subfigure[]{\includegraphics[trim=0 5 0 5,clip,height=60mm]{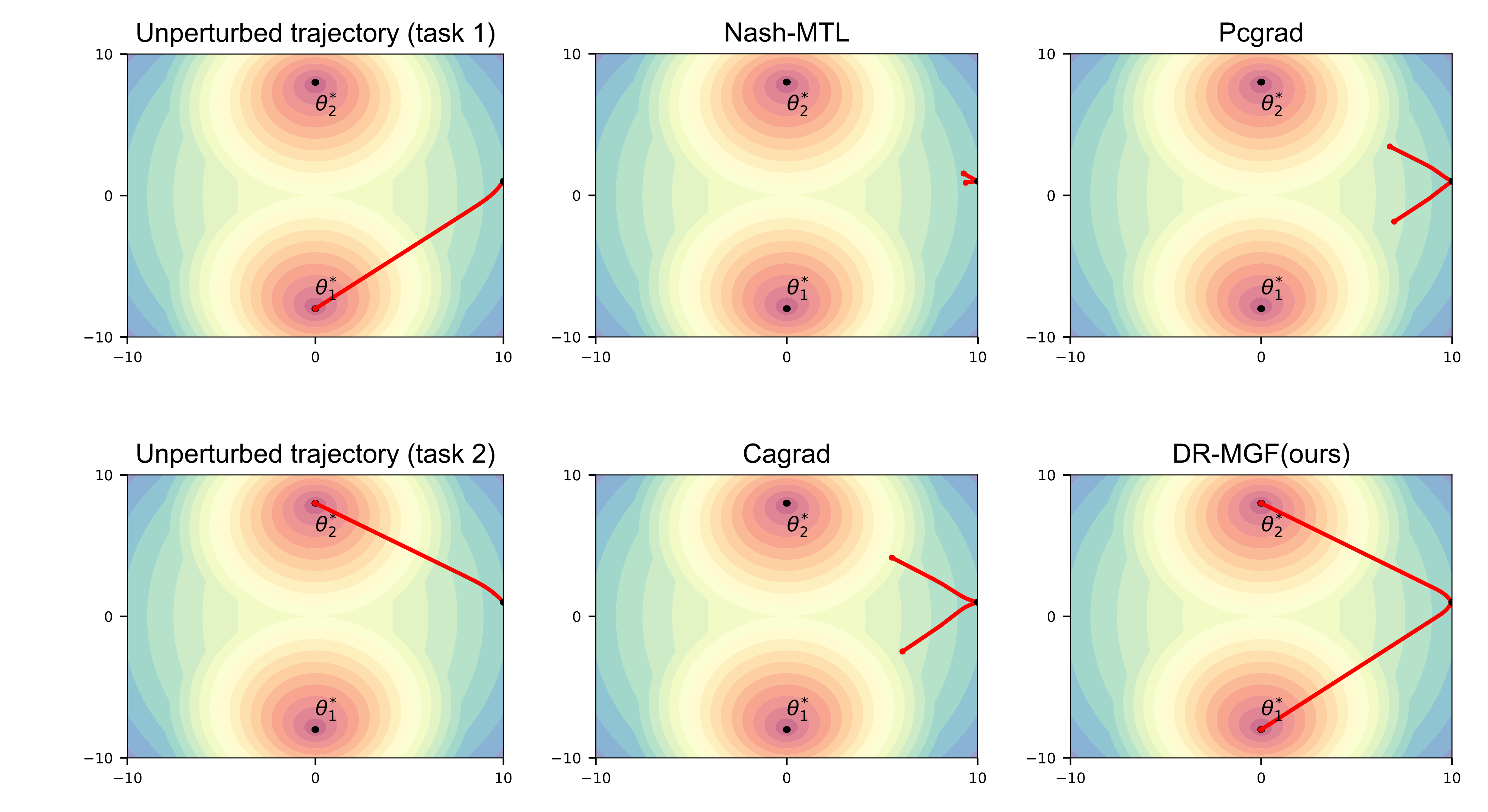}}
\caption{DR-MGF ensures that the designed two-task model converges to the optimal solution for two conflicting tasks. (a) A two-task neural network $M$, of which both task output branches:$\{\theta_{1},\theta_{2}\}$ share two parameters: $\{w_{1},w_{2}\}$, and the multi-task objective is $f_{1}(\theta_{1})+f_{2}(\theta_{2})$; (b) Compared with existing methods such as Pcgrad\cite{yu2020gradient}, Cagrad\cite{liu2021conflict} and Nash-MTL\cite{navon2022multi}, the proposed DR-MGF achieves the best performance. Please see Appendix \ref{appendix_toy} for implementation details.} 
\label{overview}
\end{figure*}
We conduct a toy experiment as shown in Fig. \ref{overview}, where we design a conflict optimization problem for a two-task model. Advanced gradient de-conflict methods such as Pcgrad\cite{yu2020gradient}, Cagrad\cite{liu2021conflict} and Nash-MTL\cite{navon2022multi} fail to converge to optimal solutions. In contrast, DR-MGF enables the model to converge to optimal points for both tasks. Furthermore, we apply the proposed method to a series of MONs such as SDN\cite{kaya2019shallow}, MSDnet\cite{huang2017multi} and Segnet-MTAN\cite{liu2019end}. Experimental results on public datasets including CIFAR, ImageNet and NYUv2 demonstrate that DR-MGF outperforms existing methods. The main contributions are as follows:
\begin{itemize}
\item  By extensive experiments, we find that tasks in MONs usually have their preferred inference routes even trained by vanilla stochastic gradient descent algorithms, but these routes are entangled. This observation motivates our work.

\item  We propose a novel gradient de-conflict method named DR-MGF to alleviate the gradient conflict problem of MONs. To the best of our knowledge, this is the first work that explores gradient de-conflict solutions from the perspective of network disentanglement, and doesn't require to modify the overall network architecture.

\item DR-MGF can be applied to enhance the performance of both multi-task networks and multi-exit networks (MONs). It achieves state-of-the-art performance on three public datasets, and can be extended to general networks with multiple learning objectives. 
\end{itemize}

This work is a significant extension of our previous conference research\cite{sun2022meta}. Firstly, compared with the previous work\cite{sun2022meta} that only contains a fusion stage, we design a novel disentanglement-and-fusion paradigm in DR-MGF, which significantly boosts the overall performance of MONs by jointly learning task-specific importance variables. Secondly, we further extend the problem of gradient de-conflict from multi-exit neural networks to general MONs. The experimental results reported in this work demonstrate the superiority of DR-MGF in addressing various types of inter-task interference. Finally, we conduct more systematic experimental studies to investigate task-specific substructures in MONs, and verify our hypotheses regarding the relationship between gradient conflicts and model convergence. The organization of this paper is as follows: we review related works in Sec. \ref{related}; then we investigate the gradient conflict problem and introduce DR-MGF in Sec. \ref{method}; the experimental results and analysis are reported in Sec. \ref{Exp}. The code will be released at \url{https://github.com/SYVAE/DR-MGF}.

\section{Related works}
\label{related}
We firstly review the gradient de-conflict methods for MONs. Then, we introduce prior research regarding the sparsity of deep neural networks, given that structural sparsity is closely related to the disentanglement and substructure extraction within neural networks.
\subsection{Gradient de-conflict methods for MONs}
This subsection reviews related works associated with the optimization methods for two types of MONs: multi-exit networks\cite{kaya2019shallow,passalis2020efficient,jie2019anytime} for depth-adaptive inference, and  multi-task networks\cite{liu2019end} for multiple-task prediction.
\subsubsection{Gradient de-conflict methods for multi-exit networks}
Depth-adaptive deep neural networks\cite{scardapane2020should} can conditionally adjust their inference depth based on the complexity of inputs. The multi-exit structure is commonly chosen to construct inference depth-adaptive networks, which attaches different output exits at various depths of the model \cite{kaya2019shallow,jie2019anytime,fei2022deecap}. By designing early-exiting policies such as the confidence-based criterion\cite{kaya2019shallow,xin2020deebert,schwartz2020right,huang2017multi,wolczyk2021zero} or the learned policy networks\cite{jie2019anytime,huang2017learning,liu2021faster} for evaluating the complexity of inputs, multi-exit networks can adaptively select the output exits and thus adjust the inference depth.\par

Yet, exits in multi-exit networks usually interfere with each other. To address this problem, some recent works\cite{phuong2019distillation,wang2021harmonized,liu2020metadistiller} proposed training algorithms based on knowledge distillation to make their learning objectives consistent. Differently, gradient-adjustment approaches, such as Pcgrad \cite{yu2020gradient,wang2020gradient}, were applied to reduce conflicts among different exits by performing gradient re-projection policies. Gradient equilibrium\cite{li2019improved} or loss weighting \cite{duggal2020elf} is applied to adjust the gradient scale of each exit for better model performance.\par 

\subsubsection{Gradient de-conflict methods for multi-task networks}
The multi-task networks have multiple output branches, with each output branch typically responsible for different tasks. If the gradients of different task objectives are not well-aligned\cite{liu2021conflict}, the joint gradients may not provide a clear convergence direction for the multi-task networks. To tackle this issue, various gradient adjustment approaches have been proposed\cite{chen2018gradnorm,liu2021towards,guo2018dynamic,liu2021conflict,javaloy2021rotograd}. \par
\textit{Magnitude-adjustment approaches} adjust the gradient magnitudes of tasks by weighting the task loss functions\cite{liuauto} or directly manipulating the gradients\cite{chen2018gradnorm,liu2021towards,yang2022adatask}. For instance, AdaTask\cite{yang2022adatask}, GradNorm\cite{chen2018gradnorm} and IMTL\cite{liu2021towards}, aim to balance the gradient magnitudes of different tasks. In \cite{guo2018dynamic}, the authors propose an adaptive loss-weighting policy to prioritize more difficult tasks. The auto-$\lambda$\cite{liuauto} employs a bi-level optimization strategy to dynamically adjust the weights of auxiliary and primary task objectives.\par

\textit{Direction-projection} approaches\cite{navon2022multi,sener2018multi,liu2021conflict,yu2020gradient} aim to find an optimization direction that can alleviate inter-task interference. For instance, Nash-MTL\cite{navon2022multi} and Cagrad\cite{liu2021conflict} optimize the overall objective of multi-task models to find a Pareto-optimal solution. In contrast, Pcgrad\cite{yu2020gradient} directly projects each conflicting gradient onto the normal plane of the other to suppress interfering components. Several methods\cite{guo2020learning,rosenbaum2018routing,vandenhende2019branched} have been proposed to enhance the multi-task learning performance by designing better network structures, and these works are beyond the scope of our work.\par

Existing gradient de-conflict methods for multi-exit and  multi-task networks treat the shared network part as a whole. Therefore, they attempt to achieve no-conflict training on all shared filters. However, works proposed in \cite{sun2020adashare,guo2020learning} suggest that  shared parameters might not be equally important to each task. We further verify this assumption in this work. Motivated by these related works and our own observations, the proposed DR-MGF starts from a new perspective of network disentanglement to address inter-task interferences. In contrast to \cite{sun2020adashare,guo2020learning,rosenbaum2018routing,vandenhende2019branched}, DR-MGF focuses on improving optimization performance without modifying the network structures. 

\subsection{Research on the sparsity of deep neural networks}
Sparsity plays an important role in scaling biological brains. The more neurons a brain has, the sparser it becomes\cite{herculano2010connectivity}. Similarly, the Lottery Ticket Hypothesis\cite{frankle2018lottery} points out that sparse structures also exist in DNNs.\par 
\subsubsection{Neural network pruning}
In \cite{meng2020pruning,hoefler2021sparsity,frankle2018lottery}, magnitude-based methods are adopted to prune dense models and obtain sparse inference substructures. For example, RigL\cite{evci2020rigging} iteratively prunes networks by selecting parameters with the top-K magnitudes, and adds new connections by selecting routes with the top-K gradient magnitudes. These works verify that magnitude is an effective measurement of filter importance when identifying model-preferred inference routes.\par
\subsubsection{Neural network disentanglement}
The superior performance of DNNs is rooted in their complex architectures and huge numbers of filters, which thereby restrict the interpretation of their internal working mechanisms\cite{li2020interpretable}. Several works\cite{hu2021architecture,li2020interpretable} have been proposed to explain the learned semantic concepts by identifying the inference substructures of the networks. For example, NAD\cite{hu2021architecture} employs information bottleneck to disentangle the inference sub-routes of different classes. \par
The sparsity of DNNs indicates three important conclusions that support our work: 1) the model parameters are not equally important to model performance, i.e., the task performance often relies on a certain sparse substructures of the model; 2) the magnitudes of filters can serve as an effective measure to evaluate their importance to different tasks; 3) DNNs are typically over-parameterized and thus possess sufficient capacity for learning multiple objectives. 

\begin{figure}[h]
\centering
\subfigure[]{\includegraphics[width=0.4\textwidth]{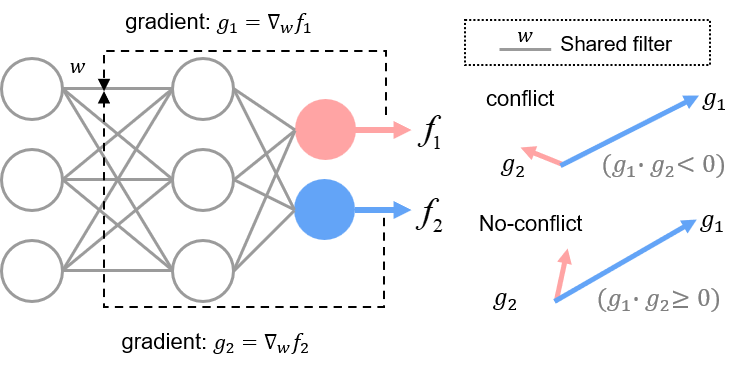}}\\
\subfigure[]{\includegraphics[trim=0 5 0 5,clip,height=0.2\textwidth]{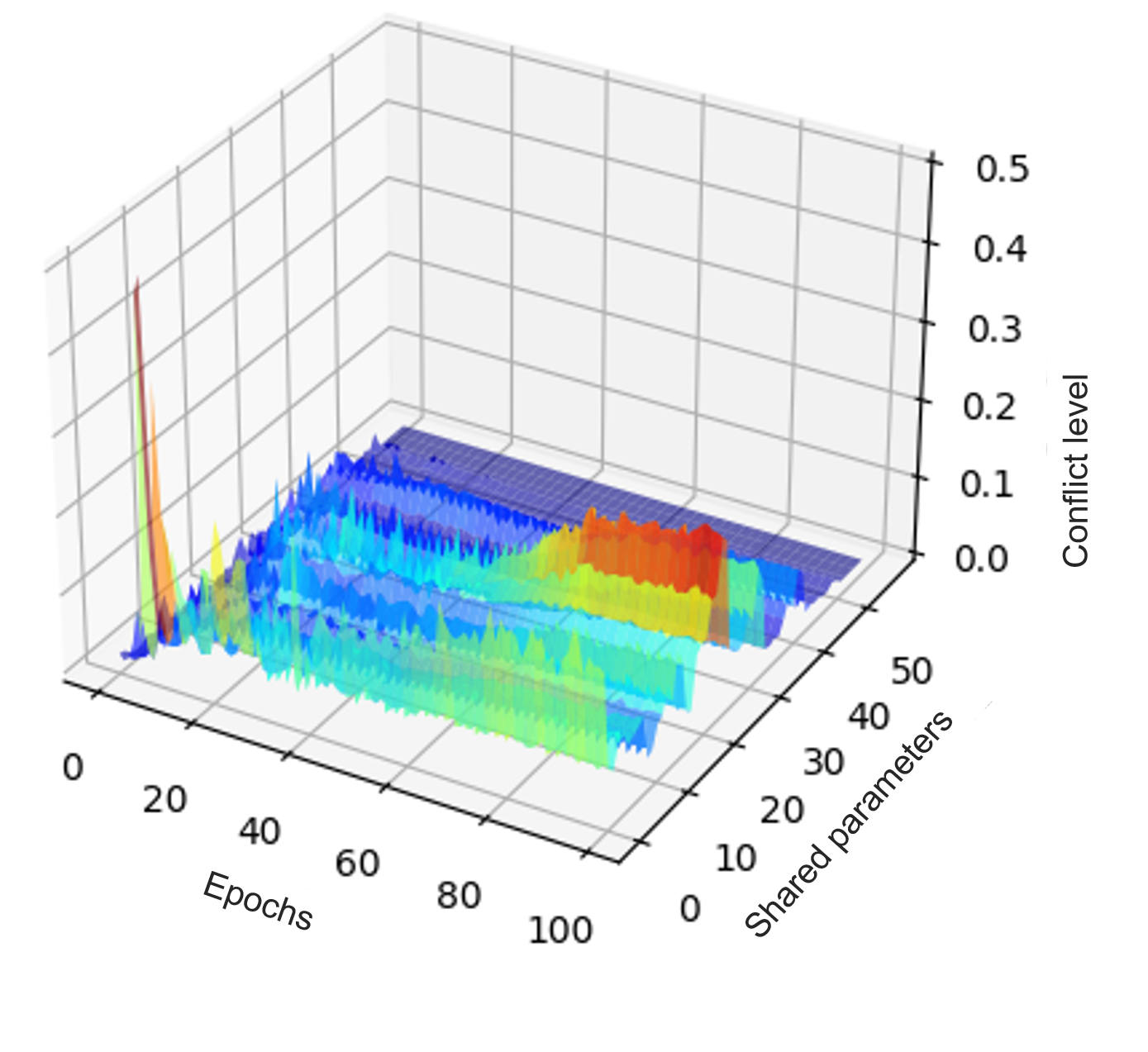}}
\subfigure[]{\includegraphics[trim=0 5 0 5,clip,height=0.2\textwidth]{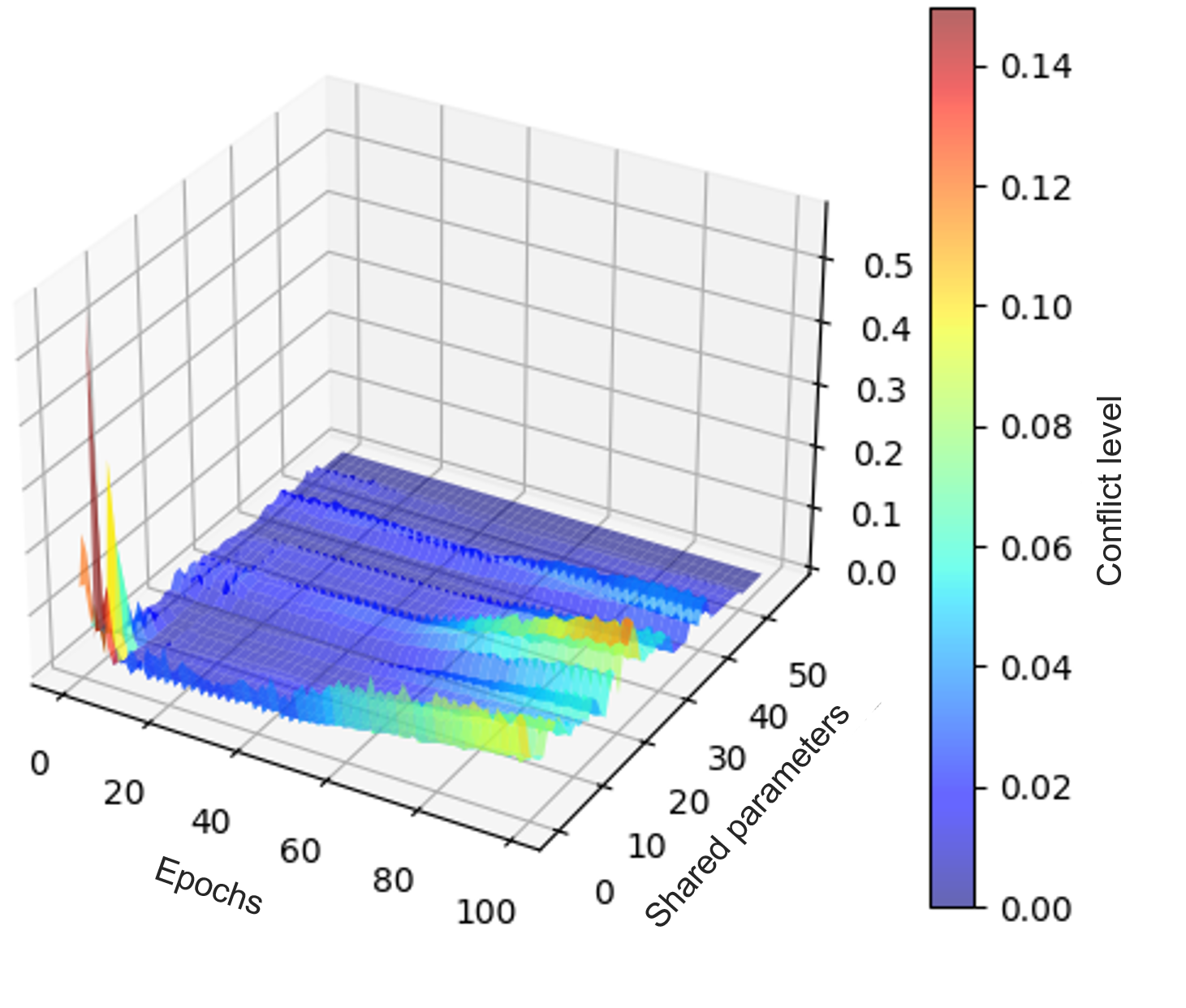}}
\caption{Illustration of the gradient conflict problem and results when applying the proposed gradient de-conflict methods. (a) Gradient conflict occurs when the cosine similarity between the gradients of two tasks is negative; (b) The gradient conflict level (please refer to Eq. (\ref{conflictlevel})) on shared filters when applying vanilla Stochastic Gradient Descent algorithm to Vgg-SDN\cite{kaya2019shallow}; (c) The gradient conflict level on shared filters when using DR-MGF.}
\label{conflictproblem}
\end{figure}

\section{Method}
\label{method}

We first make analysis of the gradient conflict problem in MONs in Sec. \ref{analysisMONs}, and then verify the existence of task-preferred substructures in MONs (Sec. \ref{substructure}). Finally, we propose a gradient de-conflict algorithm named DR-MGF (Dynamic Routes and Meta-weighted Gradient Fusion) in Sec. \ref{introduceDRMGF}.
\subsection{Analysis of the gradient conflict problem in MONs}
\label{analysisMONs}
Without loss of generality, we formulate the gradient conflict problems based on a two-output network, which has two output branches (see Fig. \ref{conflictproblem}). Given the shared filter $w$, let $g_{1}=\nabla_{w}f_{1}$ denote the gradient of $f_{1}$ with respect to $w$, and $g_{2}=\nabla_{w}f_{2}$ denote the gradient of $f_{2}$ with respect to $w$. The $f_{i}$ is the loss function for the $i$-th task.
\subsubsection{The mathematical formulation of gradient conflict} 
We take the conflict between $g_{1}$ and $g_{2}$ as an example when $g_{1}\cdot g_{2}<0$\cite{yu2020gradient}. The loss degradation of $f_{1}$ can be approximated by a first-order Taylor expansion\cite{yu2020gradient,sankararaman2020impact} as shown in Eq. (\ref{lossdegradation}) when the learning rate $\epsilon$ is small:
\begin{equation}
\centering
\label{lossdegradation}
\begin{array}{c}
\vspace{6pt}
\Delta f_{1}^{g_{1}+g_{2}}\approx -\epsilon \left(g_{1}^{2}+g_{1}g_{2}\right)+o(\epsilon^{2})\\
< -\epsilon \left(g_{1}^{2}\right)+o(\epsilon^{2})\approx \Delta f_{1}^{g_{1}}
\end{array}
,\qquad g_{1}\cdot g_{2}<0.
\end{equation}\par
As shown in Eq. (\ref{lossdegradation}), the convergence of the first output branch is negatively influenced by the conflicting gradient $g_{2}$, indicating that gradient conflict harms the convergence of MONs. Please see Appendix \ref{appendix_conflict} for detailed explanation of Eq. (\ref{lossdegradation}). \par

\subsubsection{The existence of gradient conflict and its interference on the convergence of MONs} 
\begin{figure}[h]
\centering
\subfigure[MSDnet]{\includegraphics[scale=0.28,trim=14 14 14 14,clip]{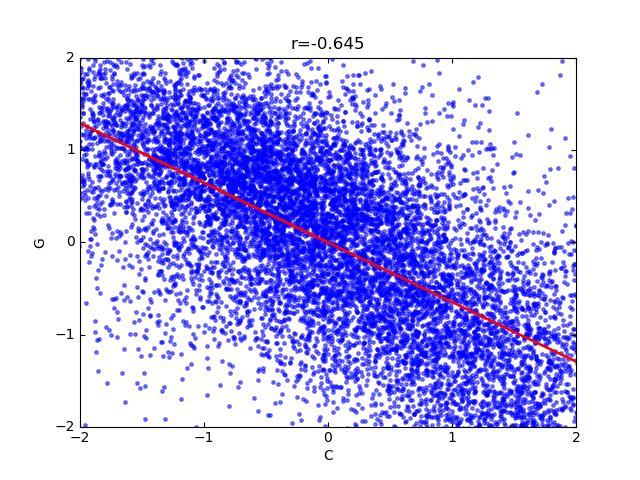}}
\subfigure[Resnet-SDN]{\includegraphics[scale=0.28,trim=14 14 14 14,clip]{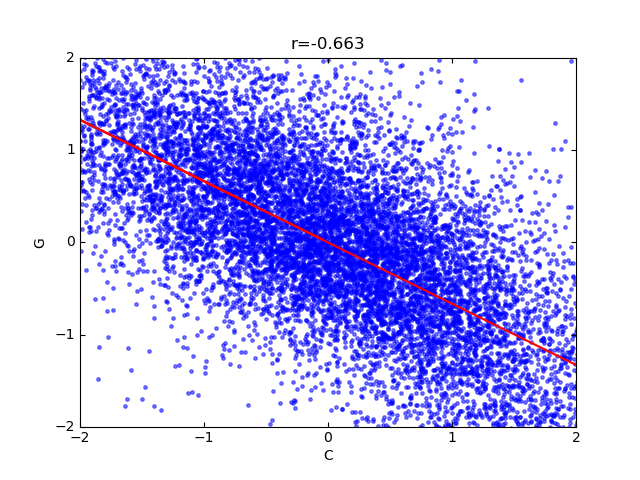}}\\
\subfigure[Vgg-SDN]{\includegraphics[scale=0.28,trim=14 14 14 14,clip]{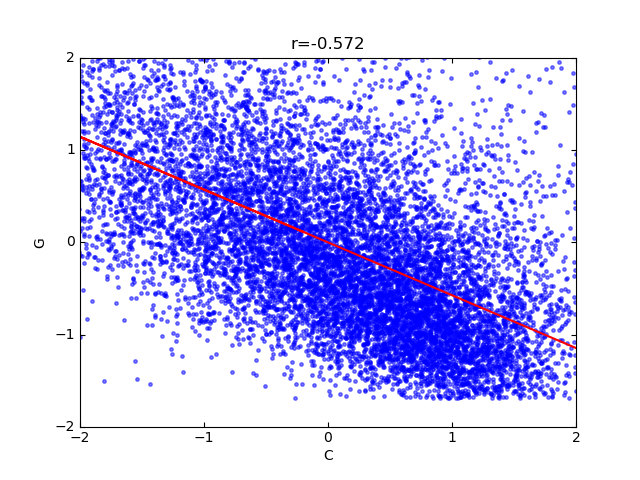}}
\subfigure[Wideresnet-SDN]{\includegraphics[scale=0.28,trim=14 14 14 14,clip]{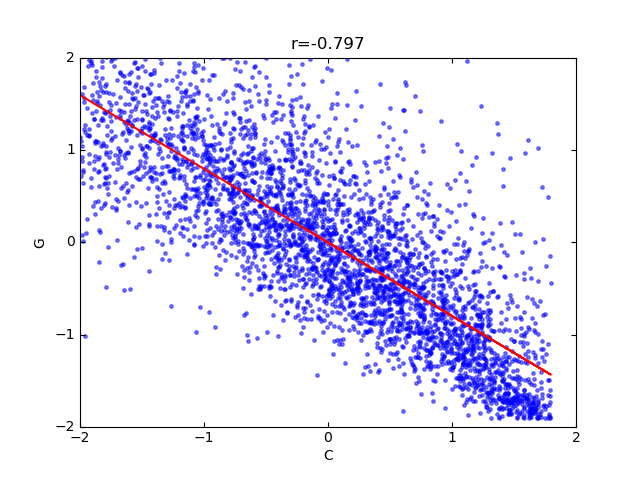}}
\caption{Pearson coefficients between the normalized gradient conflict values (C) and the normalized convergence gains (G).}
\label{Pearson}
\end{figure}

The above formulation is based on a simplified setting. To verify the existence of gradient conflict in practical DNNs and investigate its correlation with model convergence,
we conduct experiments on several popular MONs using the CIFAR100 dataset. The results are shown in Fig. \ref{Pearson}. Denoting the number of shared filters as $N$, the conflict value between two task gradients ($g_{1}$,$g_{2}$) can be calculated using Eq. (\ref{conflictlevel}). Please see Appendix \ref{appendix_relation} for detailed explanation.
\begin{align}
\centering
C=\sum_{i=1}^{N}max(0, \dfrac{-g_{1,i}^{T}g_{2,i}}{\|g_{1}\|^{2}}),
\label{conflictlevel}
\end{align}
where $\|g_{1}\|=\sqrt{\sum_{i=1}^{N}{g_{1,i}^{T}g_{1,i}}}$. The relative convergence gain of Task 1 is obtained through updating the networks by $g_{1}$ and $g_{1}+g_{2}$ as shown in Eq. (\ref{convergencegains}):
\begin{equation}
\label{convergencegains}
G=\dfrac{\Delta f_{1}^{g_{1}+g_{2}}-\Delta f_{1}^{g_{1}}}{\Delta f_{1}^{g_{1}}}.
\end{equation}\par


We calculate the Pearson coefficients between $G$ and $C$ for MSDnet\cite{huang2017multi}, and a series of SDN networks\cite{kaya2019shallow} (Resnet-SDN,Vgg-SDN, Wideresnet-SDN). The results in Fig. \ref{Pearson} indicate that the gradient conflict among tasks not only exists but also hinders the convergence of each output branch.

\begin{figure}[h]
\centering
\subfigure[]{\includegraphics[width=0.22\textwidth,trim=15 5 30 15,clip]{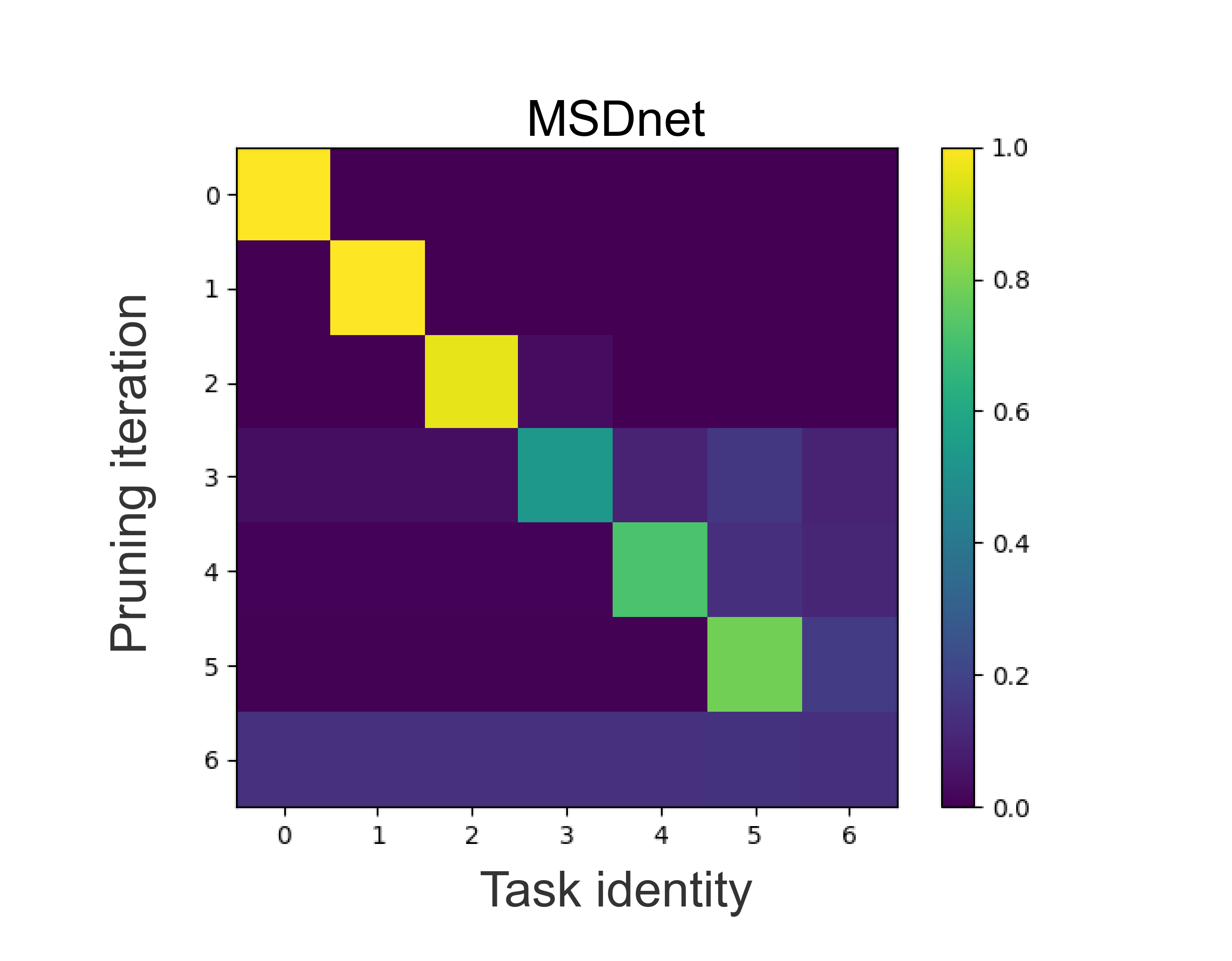}}
\subfigure[]{\includegraphics[width=0.22\textwidth,trim=15 5 30 15,clip]{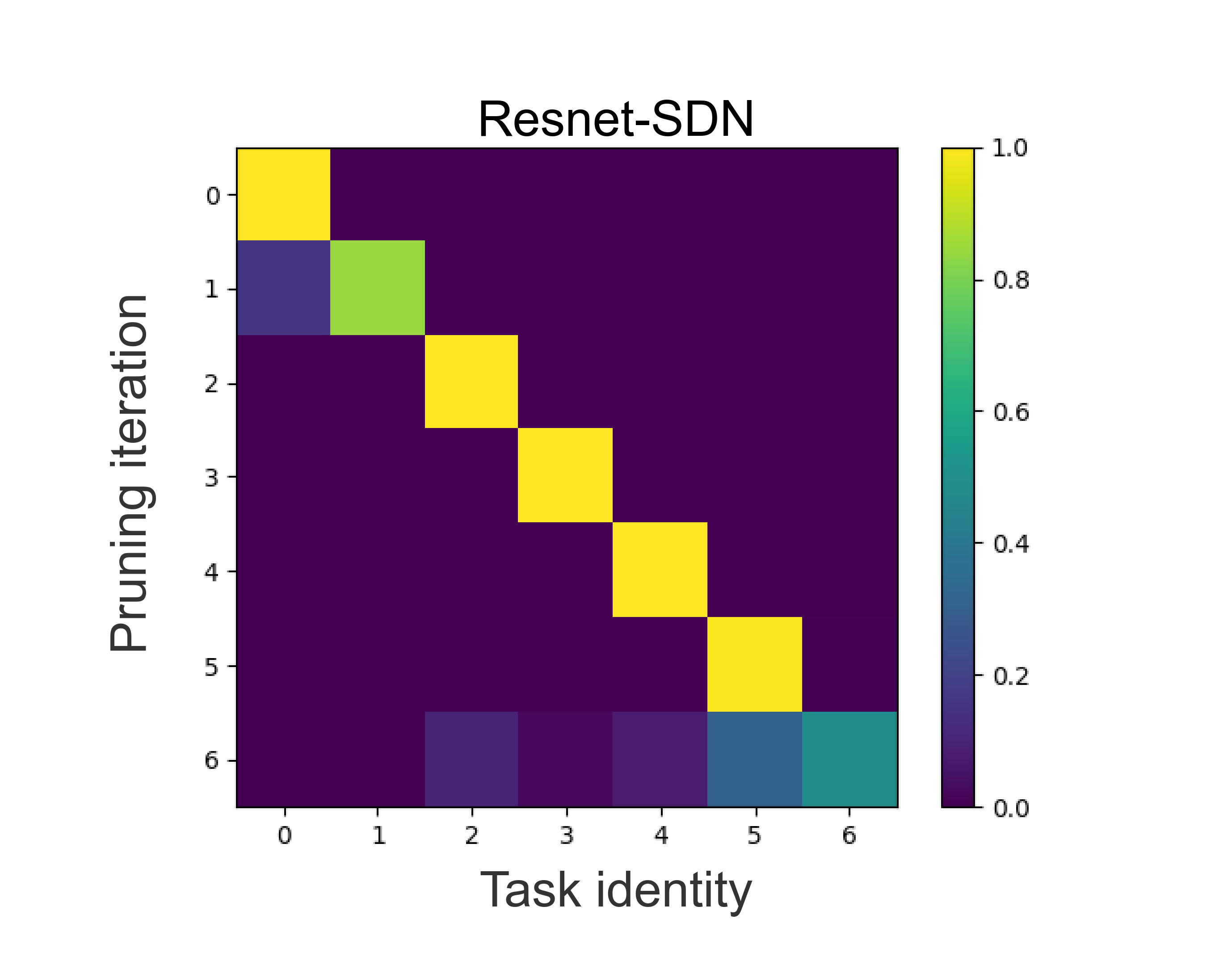}}
\subfigure[]{\includegraphics[width=0.22\textwidth,trim=15 5 30 15,clip]{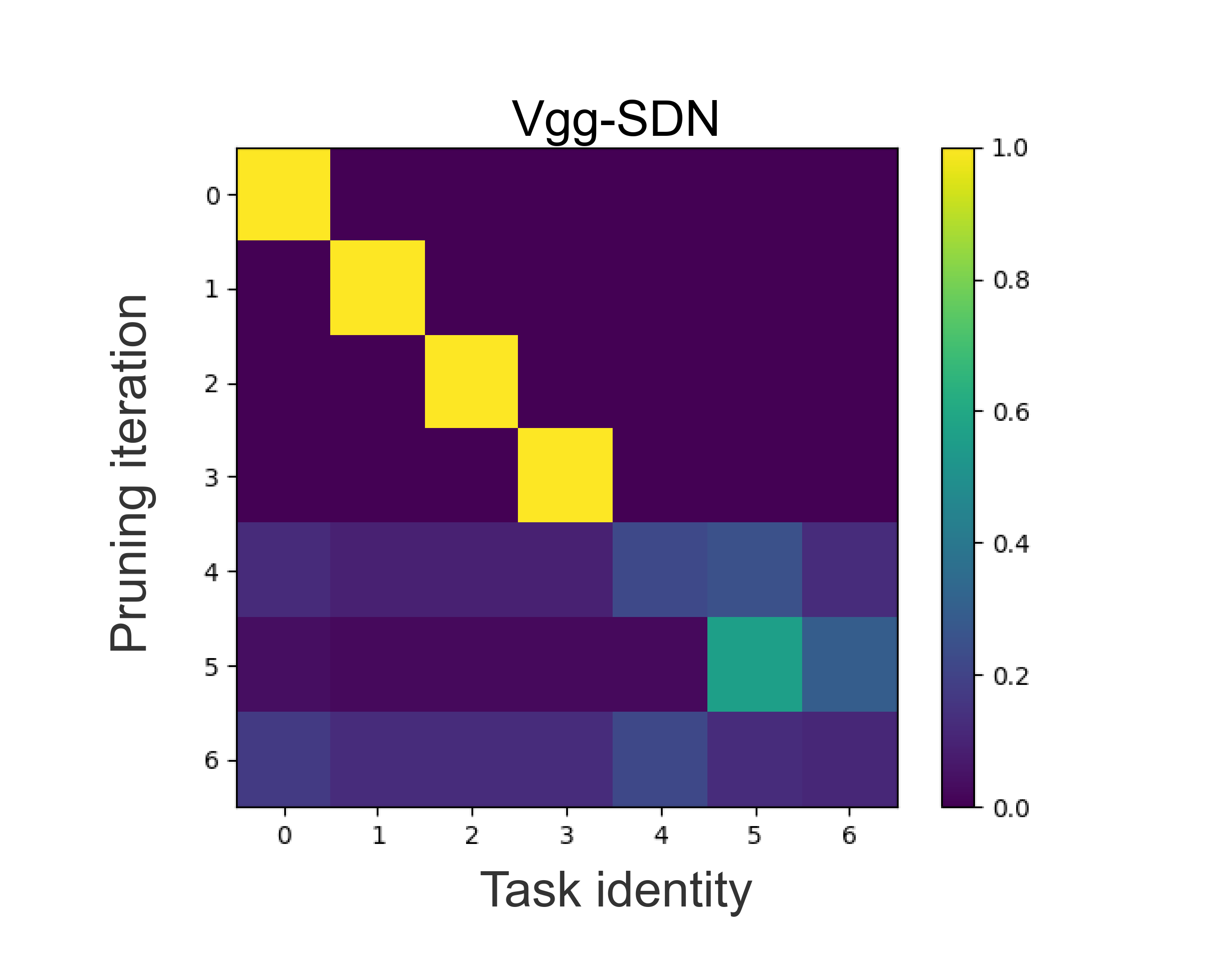}}
\subfigure[]{\includegraphics[width=0.22\textwidth,trim=15 5 30 15,clip]{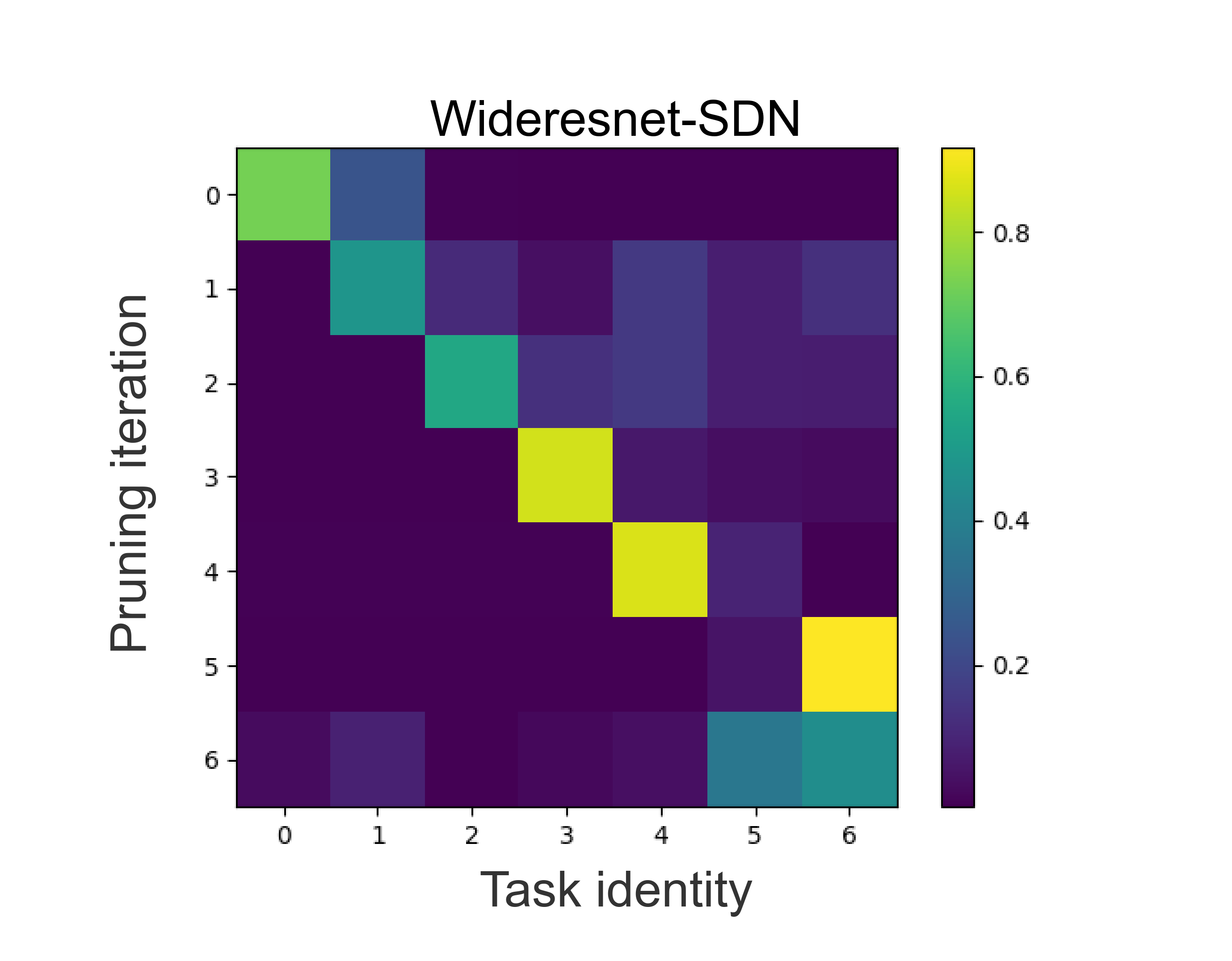}}
\caption{Relative performance degradation of different tasks when pruning the shared filters according to their importance to each task. From the top row to the bottom of each figure, we sequentially prune the important filters for each task.}
\label{degradation}
\end{figure}

\subsection{Verification and analysis of task-preferred substructures in MONs}
\label{substructure}
Different tasks share a large number of filters, which results in entanglement among tasks. We assume that these shared filters in MONs are not equally important for different tasks, which forms the basis of this work. To verify this assumption, we investigate how much the accuracy of each task degrades when pruning different filter groups. We select filters that might be more significant to a certain task based on their accumulated gradient norms of that task. Specifically, we first accumulate the normalized task-specific gradient norms of the filters throughout the entire training stage:
\begin{equation}
\centering
\label{gradnorm}
\nu_{k,i}=\dfrac{(\sum_{t=1}^{T}\|g_{k,i}^{t}\|)^{2}}{\sum_{i=1}^{N}(\sum_{t=1}^{T}\|g_{k,i}^{t}\|)^{2}},\qquad k\in [1,K],
\end{equation}
where $t$ represents the training iterations, $i$ means the index of filters, $k$ denotes the task index and $\nu_{k,i}$ represents the importance of the $i$-th filter to the $k$-th task. The normalization in Eq. (\ref{gradnorm}) alleviates the effects of gradient magnitude variation. Then, for a given task $1$, we prune the shared filters\footnote{The "prune" operation is conducted by setting the corresponding parameters to zero.} that are more important to this task than any other tasks, i.e. $\{0\cdot w_{i}|\nu_{1,i}>\nu_{k,i},k\in[2,K]\}$. It is expected to observe that the performance deterioration of task 1 will be greater than that of other tasks if our hypothesis holds. \par
As shown in Fig. \ref{degradation}, we iteratively prune the important filters of each task from the $1$st to the $7$th task. The relative accuracy degradation is plotted along the horizontal axis. It is observed that pruning the important filters of a specific task primarily decreases the performance of that task, showing that the importance of filters in MONs varies across tasks. Therefore, applying a soft-disentanglement policy to the shared model parts may help mitigate inter-task interference by allowing tasks to dominate the optimization of their preferred filters.

\subsection{Dynamic Routes and Meta-weighted Gradient Fusion}
\label{introduceDRMGF}
Motivated by the above experimental findings, we propose a novel gradient de-conflict algorithm named DR-MGF, which follows a disentanglement-and-fusion paradigm in each training epoch as shown in Fig. \ref{fig1}. In the disentanglement phase, DR-MGF simultaneously evaluates the task-specific importance of filters and calculates gradients for each task independently. Then, in the fusion stage, DR-MGF utilizes a meta-weighted gradient fusion policy to synthesize the gradients of all tasks based on the task-specific importance variables.
The main steps of the proposed DR-MGF are illustrated in \textbf{Algorithm} \ref{meta-GF}.
\subsubsection{Disentanglement of network structure}
\label{fuzzyroutes}
\begin{figure}[h]
\includegraphics[width=\linewidth]{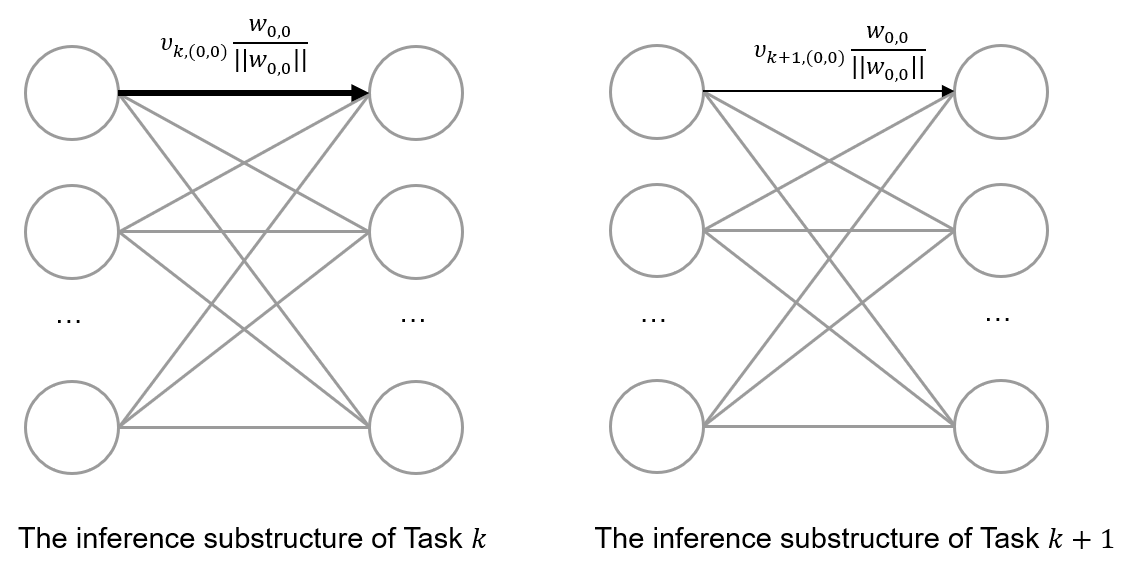}
\caption{Task-preferred inference substructures shaped by the learned task-specific importance variables between two cascaded layers. The thickness of lines indicate the magnitude values of the filter.}
\label{connection}
\end{figure}

Let $\mathcal{X}$ denote the training dataset, $\mathcal{Y}$ denote the annotations, and $f$ represent the objective function. To facilitate better understanding, we illustrate the disentanglement stage by focusing on a case study of one network layer. Denoting the filter group of this layer as $w\in \mathbb{R}^{m\times n \times s\times s}$, where $m$ and $n$ represent the numbers of output and input neurons, respectively, and $s$ is the size of filters. The topology structure is shown in Fig. \ref{connection}, with the thickness of the lines indicating the magnitude value of the filter. \par
To evaluate the importance of filters to different tasks during training, DR-MGF combines task-specific importance variables $\{\nu_{k}\}_{k\in[1,K]}$ with the normalized filters using weight normalization. As a result, each task can be optimized while learning its own inference structure without entanglement. The forward inference of the $k$-th output branch is formulated as Eq. (\ref{WN}):
\begin{equation}
\label{WN}
\centering
\begin{array}{c}
\hat{\mathcal{X}}=\Phi(\nu_{k}\dfrac{w}{\|w\|}, \mathcal{X}),\qquad \nu_{k}\in \mathbb{R}^{m\times n},
\end{array}
\end{equation}
where $k$ means the task index, "$\Phi$" stands for a specific type of operation such as convolution. The normalization operation $\frac{w}{\|w\|}$ is conducted across the $s\times s$ coordinate systems. At the end of the training, $\{\nu_{k}\}_{k\in[1,K]}$ can actually shape task-preferred inference sub-structures (see Fig. \ref{pruning}(b)).\par 

To enhance the sparsity of task-preferred inference sub-structures, we incorporate a lateral normalization operation on $\nu_{k}$ across different channels and calculate the weight-decay loss at the same time. The computational equations are given by: 
\begin{equation}
\label{sparsity}
\centering
\begin{array}{c}
\vspace{6pt}
\widehat{\nu_{k,(i,j)}}=\dfrac{\nu_{k,(i,j)}}{(\epsilon+0.1\sum_{j=1}^{n}\nu_{k,(i,j)})^{0.5}},\\
l_{k}=\sum_{i=1}^{m}\sum_{j=1}^{n}(\nu_{k,(i,j)})^{2}.
\end{array}
\end{equation}\par
Then, the optimization of the $k$-th task is formulated as Eq. (\ref{disentangleeq}):
\begin{equation}
\label{disentangleeq}
\centering
\nu_{k}^{*},w_{k}^{*}=\mathop{\arg\min}\limits_{w,\nu_{k}} f_{k}(\Phi(\widehat{\nu_{k}}\dfrac{w}{\|w\|}, \mathcal{X}), \mathcal{Y})+\lambda l_{k},
\end{equation}
where $\lambda$ is a predefined loss weight ($\lambda=0.0001$). The independent training process of each task shares the same initial weights $w_{0}$, and performs one training epoch on the dataset.

\subsubsection{Meta-weighted gradient fusion} 
After the disentanglement stage, DR-MGF employs a meta-weighted gradient fusion policy to integrate the task gradients based on the learned task-specific importance variables $\{\nu_{k}\}_{k\in [1,K]}$. This is a bi-level optimization process, which consists of an inner optimization process (InnerOpt) and an outer optimization process (OuterOpt) as described below:
\begin{equation}
\centering
\label{fusiongradient}
\begin{array}{c}
\vspace{6pt}
\nu_{k}^{'}=\dfrac{\nu_{k}}{\sum_{i=1}^{n}\nu_{k,i}}, \\
\vspace{6pt}
g=\dfrac{\sum_{k=1}^{K}\nu_{k}^{'}g_{k}}{\sum_{k=1}^{K}\nu_{k}^{'}},\\
\nu=\underbrace{\mathop{\arg\min}\limits_{\nu}\sum_{k=1}^{K}f_{k}(\Phi(\underbrace{w_{0}- g}_{\text{InnerOpt}}, \mathcal{X}), \mathcal{Y})}_{\text{OuterOpt}}.
\end{array}
\end{equation}\par
In the inner optimization, DR-MGF utilizes a differentiable update module to optimize $w_{0}$ through joint gradient $g$. In the outer optimization, DR-MGF optimizes $\{\nu_{k}\}_{k\in [1,K]}$ based on the computed task objectives. It is worth noting that the gradients $\{g_{k}\}_{k\in [1,K]}$ in this stage represent expected updating direction of each task for the filter, and the computation is formulated as:
\begin{align}
\label{expectedgradient}
g_{k}&=\widehat{\nu_{k}^{*}}\dfrac{w^{*}_{k}}{\|w^{*}_{k}\|}-w_{0} \qquad g_{k}\in \mathbb{R}^{m\times n \times s \times s}.
\end{align}\par
Specifically, the expected gradient $g_{k}$ of the $k$-th task is obtained after independently training this task for one epoch as shown in Eq. (\ref{disentangleeq}).
\par

\floatname{algorithm}{Algorithm}
\renewcommand{\algorithmicrequire}{\textbf{Input:}}
\renewcommand{\algorithmicensure}{\textbf{Output:}}
\begin{algorithm}
 \caption{DR-MGF (Dynamic Routes and Meta Gradient Fusion):}
    \label{meta-GF}
    \setstretch{1.1}
	\begin{algorithmic}[1]
	    \REQUIRE Initial filter weights: $w_{0}$, training dataset: $\mathcal{X}$, ground truth: $\mathcal{Y}$, task-specific importance variables: $\nu=\{\nu_{k}\}_{k\in [1,K]}$, number of tasks:$K$, objective function of each task: $F=\{f_{1},...,f_{K}\}$. The maximum training epoch number is \text{MaxIter}.
        \ENSURE  $w_{0}$
        \WHILE {$i<\text{MaxIter}$}
        	\STATE \textit{$\blacktriangledown$ 1. Disentanglement of network structure}:
        		\FOR {$k=1,...,K$}
        		   \STATE $w =w_{0}$
        		   \STATE $w,\nu_{k}=\mathop{\arg\min}\limits_{w,\nu_{k}} f(\Phi(\widehat{\nu_{k}}\dfrac{w}{\|w\|},\mathcal{X}),\mathcal{Y})+\lambda l_{k}$
        			\STATE $g_{k}=\widehat{\nu_{k}}\dfrac{w}{\|w\|}-w_{0},$
        		\ENDFOR
        	\STATE \textit{$\blacktriangledown$ 2. Meta-weighted gradient fusion}:
        	\STATE $\nu_{k}^{'}=\dfrac{\nu_{k}}{\sum_{i=1}^{n}\nu_{k,i}}, k\in [1,K]$
        	\STATE $g=\dfrac{\sum_{k=1}^{K}\nu_{k}^{'}g_{k}}{\sum_{k=1}^{K}\nu_{k}^{'}}$
        	\STATE  $\nu=\underbrace{\mathop{\arg\min}\limits_{\nu}\sum_{k=1}^{K}f_{k}(\Phi(\underbrace{w_{0}- g}_{\text{InnerOpt}}, \mathcal{X}), \mathcal{Y})}_{\text{OuterOpt}}$
        	\STATE $w_{0}=w_{0}- \dfrac{\sum_{k=1}^{K}\nu_{k}^{'}g_{k}}{\sum_{k=1}^{K}\nu_{k}^{'}}$
        \ENDWHILE        
        \RETURN $w_{0}$
    \end{algorithmic}
\end{algorithm}

\begin{remark}
\textit{\textbf{Why do we use expected gradients instead of mini-batch gradients?} When applying meta-weighted gradient fusion policy to mini-batch gradients, two potential issues arise. Firstly, in the current mini-batch training settings, the gradients produced by SGD are noisy\cite{hoefler2021sparsity,sankararaman2020impact}, and the uncertainty inherent in mini-batch gradients can negatively impact the estimation of task-specific optimization directions. Secondly, this requires to frequently change the task-preferred inference structure if employing meta-weighted gradient fusion policy during each training batch. This would make the learning process of batch normalization (BN) layers unstable, as different tasks typically have different statistics in their BN layers. An effective way to address both issues simultaneously is to fuse the expected gradient of each task as mentioned above.}
\end{remark}

\section{Experiments}
\label{Exp}
To verify the effectiveness of the proposed DR-MGF, we conduct extensive experiments on CIFAR\cite{krizhevsky2009learning}, ImageNet (ILSVRC 2012)\cite{deng2009imagenet}), and NYUv2\cite{couprie2013indoor}. Experiments on the first type of MONs (multi-exit networks) are conducted using image classification datasets (CIFAR and ImageNet), while those on the second type (multi-task networks) are performed on NYUv2. Additionally, we provide a detailed analysis of the proposed approach.

\begin{table*}[t]
\begin{minipage}[c]{1\textwidth}
\caption{Classification accuracy of individual classifiers in Vgg-SDN\cite{kaya2019shallow} on CIFAR100 and CIFAR10.}
\label{cifartable_vgg}
\resizebox{180mm}{!}{
\begin{tabular}{|rcc|ccccccc|ccccccc|}
\hline 
\multirow{2}{*}{}&\multirow{2}{*}{Params(M)}&\multirow{2}{*}{flops(M)}&\multicolumn{7}{c|}{CIFAR100}&\multicolumn{7}{c|}{CIFAR10}\\ \cline{4-17}
& &  &Vgg-SDN &GE\cite{li2019improved}& Cagrad\cite{liu2021conflict}& Pcgrad\cite{yu2020gradient}&Nash-MTL\cite{navon2022multi}&Meta-GF\cite{sun2022meta}&DR-MGF&Vgg-SDN&GE\cite{li2019improved} & Cagrad\cite{liu2021conflict}& Pcgrad\cite{yu2020gradient}&Nash-MTL\cite{navon2022multi}&Meta-GF\cite{sun2022meta}&DR-MGF \\ \hline
\rowcolor{gray!40}
Average&-&-         &66.34&66.19&67.60&66.61&      64.29      &68.51 &\textbf{69.25 }&88.15 &88.08 &89.54 & 88.05&89.00&89.82 &\textbf{90.50} \\
Exit-1& 0.05&39.76  &44.42 &44.46 &\textbf{53.08} &44.59&40.17&49.91 &51.25          &69.03 &68.97 & \textbf{76.27} &67.41&71.94&74.92 &74.67\\ 
Exit-2& 0.29&96.52  &61.08 &61.00 &61.39          &63.02&57.46&61.09 &\textbf{64.88} &84.72 &84.52 & 86.6&\textbf{88.69}&86.31&88.69&88.64 \\ 
Exit-3& 1.22&153.25 &69.80 &69.54 &70.90          &70.04&68.14&71.38 &\textbf{72.11} &92.15 &92.02 &92.40 &91.80&92.45&92.75&\textbf{93.65} \\ 
Exit-4&1.85 &191.08 &72.23 &72.11 &71.55          &73.14&69.92&\textbf{75.77}&73.89 &92.50 &92.62 & 92.79& 92.74&92.89&93.07& \textbf{94.06}\\ 
Exit-5&5.47 &247.81 &72.48 &72.32 &72.41          &72.59&71.58&74.12 &\textbf{74.37} &92.46 &92.78 &92.99 &92.75 &93.16&93.13&\textbf{94.20}\\ 
Exit-6&7.86 &285.68 &72.63 &72.38 &72.45          &72.54&71.42&74.23 &\textbf{74.41} &93.59 &92.83 &93.07 & 92.70&93.18&93.12&\textbf{94.16}\\ 
Exit-7&15.47 &314.45 &71.76&71.58 &71.43          &71.39&71.37&73.10 &\textbf{73.87} &92.61& 92.85& 93.00& 93.69&93.22&93.07&\textbf{94.15}\\ \hline 
\end{tabular}}
\end{minipage}
\end{table*}

\begin{table*}[t]
\begin{minipage}[c]{1\textwidth}
\caption{Classification accuracy of individual classifiers in Resnet-SDN\cite{kaya2019shallow} on CIFAR100 and CIFAR10.}
\label{cifartable_resnet}
\resizebox{180mm}{!}{
\begin{tabular}{|rcc|ccccccc|ccccccc|}
\hline 
\multirow{2}{*}{}&\multirow{2}{*}{Params(M)}&\multirow{2}{*}{flops(M)}&\multicolumn{7}{c|}{CIFAR100}&\multicolumn{7}{c|}{CIFAR10}\\ \cline{4-17}
& &  &Resnet-SDN &GE\cite{li2019improved}& Cagrad\cite{liu2021conflict}& Pcgrad\cite{yu2020gradient}&Nash-MTL\cite{navon2022multi}&Meta-GF\cite{sun2022meta}&DR-MGF&Resnet-SDN&GE\cite{li2019improved} & Cagrad\cite{liu2021conflict}& Pcgrad\cite{yu2020gradient}&Nash-MTL\cite{navon2022multi}&Meta-GF\cite{sun2022meta}&DR-MGF \\ \hline
\rowcolor{gray!40}
Average&- & -      &59.29&60.10&60.14&59.54&61.01&60.32&\textbf{61.84}    &86.13 &85.91 &87.04 &85.76   &87.19&86.62 &\textbf{87.33}\\
Exit-1& 0.02&19.50 &40.20&42.10&\textbf{48.73}&40.10&46.92&44.41&48.43   &71.64 &71.37 &\textbf{80.94} &69.74&77.26&76.04 &76.96 \\ 
Exit-2& 0.04&38.54 &45.45&46.91&47.05&45.67&48.74&47.17&\textbf{51.26}   &78.10 &77.11 &80.24          &77.24&\textbf{80.74}&78.11&80.21\\ 
Exit-3& 0.10&56.47 &59.08&59.85&57.77&60.04&59.68&59.70&\textbf{61.46}   &87.32 &87.21 &86.31          &87.75&87.72&86.43&\textbf{88.21} \\ 
Exit-4& 0.18&75.43 &62.40&63.81&62.62&63.47&63.39&63.25&\textbf{64.02}   &89.85 &89.63 &88.62          &89.79&89.54&89.09&\textbf{90.26}\\ 
Exit-5& 0.36&93.32 &67.88&\textbf{68.52}&67.16&67.78&68.08&68.38&68.27   &91.45 &91.51 &90.73          &\textbf{91.53}&91.30&91.48&91.49\\ 
Exit-6&0.67 &112.25&70.06&69.88&69.26&69.70&70.18&\textbf{70.25}&70.02   &92.26 &\textbf{92.33} &91.31          &92.17&91.98&\textbf{92.33}&92.05\\ 
Exit-7&0.89 &126.44&70.02&69.63&68.40         &70.07&\textbf{70.28}&70.08&69.42   &92.33 &92.21 &91.19          &92.09&91.83&\textbf{92.87}&92.17\\ \hline 
\end{tabular}}
\end{minipage}
\end{table*}

\begin{table*}[t]
\begin{minipage}[c]{1\textwidth}
\caption{Classification accuracy of individual classifiers in MSDnet on CIFAR100 and CIFAR10.}
\label{cifartable}
\resizebox{180mm}{!}{
\begin{tabular}{|rcc|ccccccc|ccccccc|}
\hline 
\multirow{2}{*}{}&\multirow{2}{*}{Params(M)}&\multirow{2}{*}{flops(M)}&\multicolumn{7}{c|}{CIFAR100}&\multicolumn{7}{c|}{CIFAR10}\\ \cline{4-17}
& &  &MSDnet &GE\cite{li2019improved}& Cagrad\cite{liu2021conflict}& Pcgrad\cite{yu2020gradient}&Nash-MTL\cite{navon2022multi}&Meta-GF\cite{sun2022meta}&DR-MGF&MSDnet&GE\cite{li2019improved} & Cagrad\cite{liu2021conflict}& Pcgrad\cite{yu2020gradient}&Nash-MTL\cite{navon2022multi}&Meta-GF\cite{sun2022meta}&DR-MGF \\ \hline
\rowcolor{gray!40}
Average&-&-&72.83&73.80&73.96&74.14&72.69&74.57&\textbf{74.62}                    &93.80&94.11&94.01 &94.09&93.12&94.40&\textbf{94.47}\\
Exit-1&0.90&56.43 &66.41 &67.74 &\textbf{68.78} &67.06&64.14&67.97&68.47         &91.13&92.02&92.19 &91.66&90.72&92.38&\textbf{92.44} \\ 
Exit-2&1.84&101.00&70.48 &71.87 &\textbf{72.55} &71.37&69.23&72.27&72.39       &92.91&93.53&93.49 &93.59&92.43&\textbf{94.22}& 94.06\\ 
Exit-3&2.80&155.31&73.25 &73.81 &74.23          &74.86&72.64&\textbf{75.06}&74.60        &93.98&94.14&94.47 &94.32&93.29&94.49&\textbf{94.77} \\ 
Exit-4&3.76&198.10&74.02 &75.13 &74.97          &75.78&74.89&75.77&\textbf{75.81}       &94.46&94.49&94.45 &94.60&93.76&\textbf{94.96}&94.81 \\ 
Exit-5&4.92&249.53&74.87 &75.86 &75.35          &76.25&75.32 &76.38&\textbf{76.72}        &94.68&94.73&94.48 &94.81&93.69&94.82&\textbf{95.02} \\ 
Exit-6&6.10&298.05&75.33 &76.23 &75.82          &76.95&75.88&\textbf{77.11}&77.01        &94.78&94.89&94.53 &94.83&93.93&94.97& \textbf{95.05}\\ 
Exit-7&7.36&340.64&75.42 &75.98 &76.08          &76.71&76.75&\textbf{77.47}&77.34      &94.64&94.96&94.48 &94.82&94.02&94.97&\textbf{95.15} \\ \hline 
\end{tabular}}
\end{minipage}
\end{table*}

\subsection{Performance on multi-exit neural networks}
\subsubsection{Datasets}
CIFAR100 and CIFAR10 both contain 60,000 RGB images. In each dataset, 50,000 images are used for training and 10,000 for testing. The images in CIFAR10 and CIFAR100 belong to 10 classes and 100 classes, respectively. We adopt the same data augmentation policy as introduced in \cite{li2019improved}. We select 5000 images from the training sets of CIFAR100 and CIFAR10 for validation. The ImageNet dataset contains 1000 classes. Its training set has 1.2 million images and we select 50,000 of them for validation. In this work, we refer to the public validation set of ImageNet as the test set because the true test set has not been made publicly available.\par

\subsubsection{Network structures}  
The classification models used in this section are two types of multi-exit networks: SDN-style networks\cite{kaya2019shallow} and MSDnet\cite{huang2017multi}. Both types of models attach several classification branches at different depths within the networks. The SDN-style networks contain two specific models: Vgg-SDN and Resnet-SDN. On the CIFAR datasets, all multi-exit networks are configured with 7 exits, and the depth of these exits are set according to the designs in \cite{kaya2019shallow,huang2017multi}. The input size of the image is $32\times 32$. On ImageNet, we validate the proposed approach using MSDnet, which is configured with 5 exits, and the input image size is set to $224\times 224$.

\subsubsection{Implementation Details}
We take the same implementation settings as introduced in \cite{kaya2019shallow, huang2017multi}. Specifically, we optimize all models using SGD with a batch size of 64 on CIFAR and 512 on ImageNet. The momentum and weight decay are set to 0.9 and $10^{-4}$, respectively. For MSDNet, we train for 300 epochs on CIFAR datasets and 90 epochs on ImageNet. For SDN-style networks, the maximum number of epochs is set to 100 on CIFAR datasets. The adjustment of the learning rate is achieved by multi-step policy: it's initially set to 0.1 and divided by 10 after $0.5\times \text{maxiter}$ and $0.75\times \text{maxiter}$ epochs.\par
We initialize the task-specific importance variables $\{\nu_{k}\}$ using KaimingInit\footnote{This is an official implementation provided by pytorch library:\url{https://pytorch.org/}.}, and optimize them using SGD with a momentum of 0.9 and a weight decay of $10^{-5}$. The initial learning rate is set to $10^{-1}$, and we take the same multi-step policy mentioned above to adjust the learning rate.\par
In the disentanglement stage, we combine task-specific importance variables with network weights, enabling each task to learn both its importance variables and expected updating directions. However, training each task independently overwhelms the inter-task cooperation. To address this, while focusing on the currently selected primary task, we simultaneously train other tasks with a small auxiliary learning rate which is lower than the primary task\footnote{The gradient magnitudes of different tasks are kept at the same scale.}. Consequently, the original objectives in Eq. (\ref{disentangleeq}) can be reformulated as follows: \par
\begin{equation}
\label{auxtraining}
\begin{array}{c}
\vspace{6pt}
\nu_{k}^{*},w^{*}=\mathop{\arg\min}\limits_{w,\nu_{k}} f_{k}(\Phi(\widehat{\nu_{k}}\dfrac{w}{\|w\|},\mathcal{X}),\mathcal{Y})\\
\vspace{6pt}
+ \sum_{i=1,i\neq k}^{K}\beta_{i} f_{aux,i}(\Phi(\widehat{\nu_{k}}\dfrac{w}{\|w\|},\mathcal{X}),\mathcal{Y})\\
+\lambda l_{k}, \,\{\beta_{i}\} \in [0, 0.5],
\end{array}
\end{equation}
where $\{\beta_{i}\}$ are empirically set to 0.4. \par

\subsubsection{Comparison of classification accuracy}
We compare the proposed DR-MGF with five representative approaches: MSDnet\cite{huang2017multi}/SDN\cite{kaya2019shallow}, GE\cite{li2019improved}, Pcgrad\cite{yu2020gradient}, Cagrad\cite{liu2021conflict}, Nash-MTL\cite{navon2022multi}. MSDnet\cite{huang2017multi}/SDN serves as baseline in the experiments, which takes the vanilla SGD as the optimizer. GE (Gradient Equilibrium) is designed to control the variance of joint gradients. Pcgrad, Cagrad, and Nash-MTL are three state-of-the-art gradient de-conflict methods proposed for multi-task networks. In this experiment, we apply them to multi-exit networks to extend the comparison.\par

As illustrated in Tables \ref{cifartable_vgg}-\ref{cifartable}, we compare the top-1 prediction accuracy of different exits when using different approaches on CIFAR datasets. On CIFAR100, existing gradient de-conflict approaches outperform the baseline. Notably, Cagrad achieves superior performance at shallow exits\cite{liu2021conflict}, yet this comes at the cost of reduced performance at deeper exits.\par
In contrast, the proposed DR-MGF aims to solve gradient conflicts among tasks from the perspective of network disentanglement. As shown in Tables \ref{cifartable_vgg}-\ref{cifartable}, DR-MGF enables multi-exit networks to achieve the best overall performance (average accuracy), on both CIFAR10 and CIFAR100 datasets. For example, the average accuracy of Resnet-SDN trained by DR-MGF is 61.84\% which surpasses the second one by $1.52$ percentage points, representing the cumulative gains of all exits achieving 10.64 ($1.52\times 7$) percentage points.\par

To further verify the effectiveness of the proposed DR-MGF, we compare existing approaches and our method on the larger ImageNet dataset, using MSDnet with 5 exits. As shown in Table \ref{imagenettable}, DR-MGF still outperforms previous approaches.\par
\begin{table}[!htb]
\centering
\caption{Classification accuracy of individual classifiers in MSDnet on ImageNet.}
\resizebox{90mm}{!}{
\begin{tabular}{|rcc|cccccc|}
\toprule
\multirow{2}{*}{}&\multirow{2}{*}{Params(M)}&\multirow{2}{*}{flops(M)}&\multicolumn{6}{c|}{ImageNet}\\ \cline{4-9}
& & & MSDnet &GE\cite{li2019improved}&  Pcgrad\cite{yu2020gradient}& Cagrad\cite{liu2021conflict}&Meta-GF\cite{sun2022meta}&DR-MGF \\ \hline
\rowcolor{gray!40}
Average&-&-&66.72&66.94 &66.98 &65.42 &67.25&\textbf{67.35}\\  
Exit-1&4.24&339.90& \textbf{58.48}&57.75 & 57.62&58.37&57.43 &58.35\\ 
Exit-2&8.77&685.46&\textbf{65.96}&65.54  & 64.87&64.21&64.82 &65.01 \\ 
Exit-3&13.07&1008.16&68.66 &\textbf{69.24}  &68.93 &66.88&69.08 &69.15 \\ 
Exit-4&16.75&1254.47&69.48 &70.27  &71.05 &68.22 &\textbf{71.67}&71.22 \\ 
Exit-5&23.96&1360.53&71.03 &71.89 &72.45 &69.42&\textbf{73.27} &73.01 \\ \bottomrule 
\end{tabular}
}
\label{imagenettable}
\end{table}

\begin{table}[h]
\centering
\caption{Classification accuracy of individual classifiers on CIFAR100 when combining DR-MGF with knowledge distillation approach.}
\resizebox{80mm}{!}{
\begin{tabular}{|rcc|ccc|}
\toprule
\multirow{2}{*}{}&\multirow{2}{*}{Params(M)}&\multirow{2}{*}{flops(M)}&\multicolumn{3}{c|}{CIFAR100}\\ \cline{4-6}
&         &       &MSDnet&KD\cite{phuong2019distillation} &KD\cite{phuong2019distillation}+DR-MGF \\  \hline
\rowcolor{gray!40}
Average&-&-       &72.83&72.97 &\textbf{74.13}\\ 
Exit-1&0.90&56.43 &66.41&66.08 &\textbf{68.01}\\ 
Exit-2&1.84&101.00&70.48&71.24  &\textbf{72.34} \\ 
Exit-3&2.80&155.31&73.25 &72.28  &\textbf{74.48}  \\ 
Exit-4&3.76&198.10&74.02 &74.50  &\textbf{75.58}  \\ 
Exit-5&4.92&249.53&74.87 &74.83 &\textbf{76.14} \\
Exit-6&6.10&298.05&75.33 &75.77 &\textbf{76.15} \\
Exit-7&7.36&340.64&75.42 &76.14 &\textbf{76.26} \\ \bottomrule 
\end{tabular}
}
\label{KD}
\end{table}

\begin{figure}[h]
    \centering 
    \includegraphics[width=0.45\textwidth]{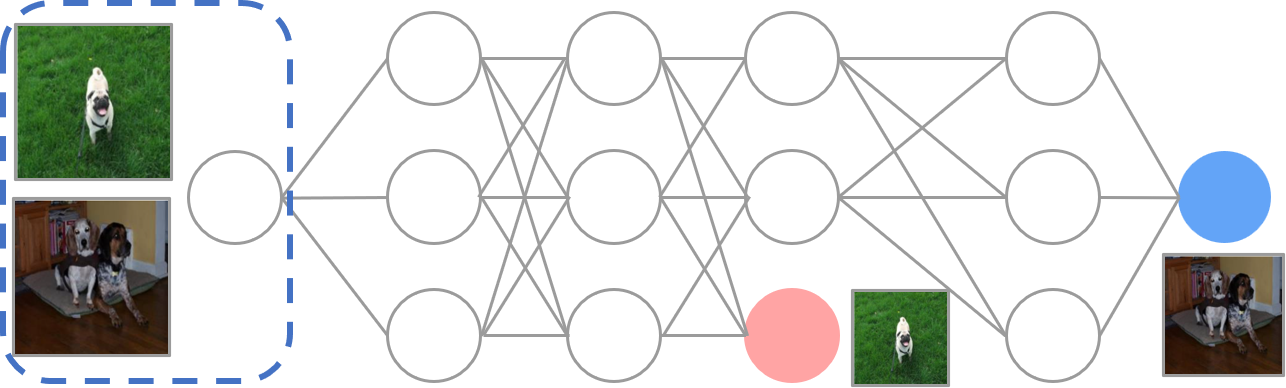} 
    \caption{Depth-adaptive inference mode of multi-exit networks. In adaptive inference mode, the multi-exit networks dynamically adjust the inference depth according to the input complexity and consequently save the computation resources.}
\label{inferenceadaptive}  
\end{figure}

\begin{figure*}[t]
\centering
\subfigure[Vgg-SDN]{
\includegraphics[height=40mm,trim=0 0 0 0,clip]{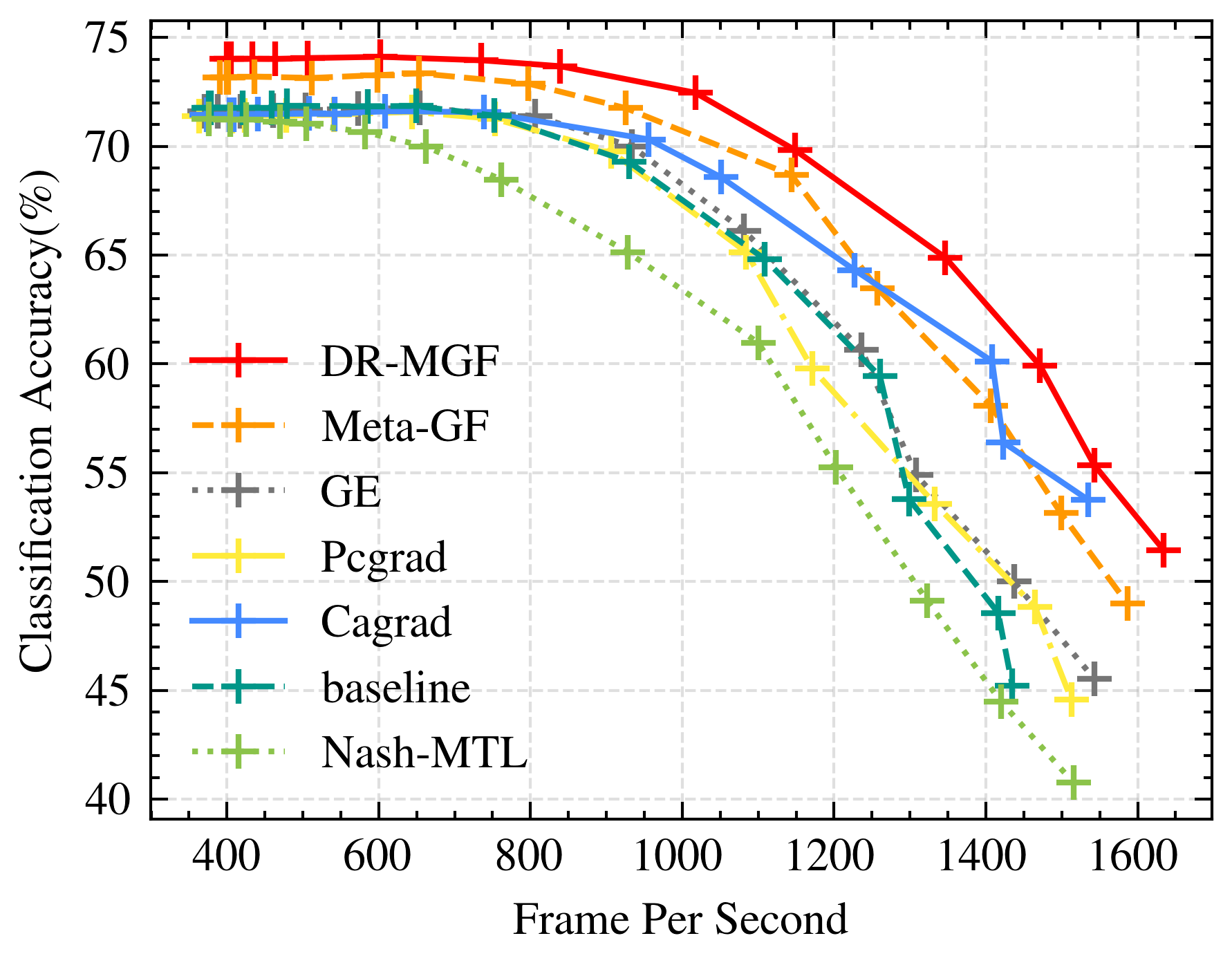}}
\subfigure[Resnet-SDN]{
\includegraphics[height=40mm,trim=0 0 0 0,clip]{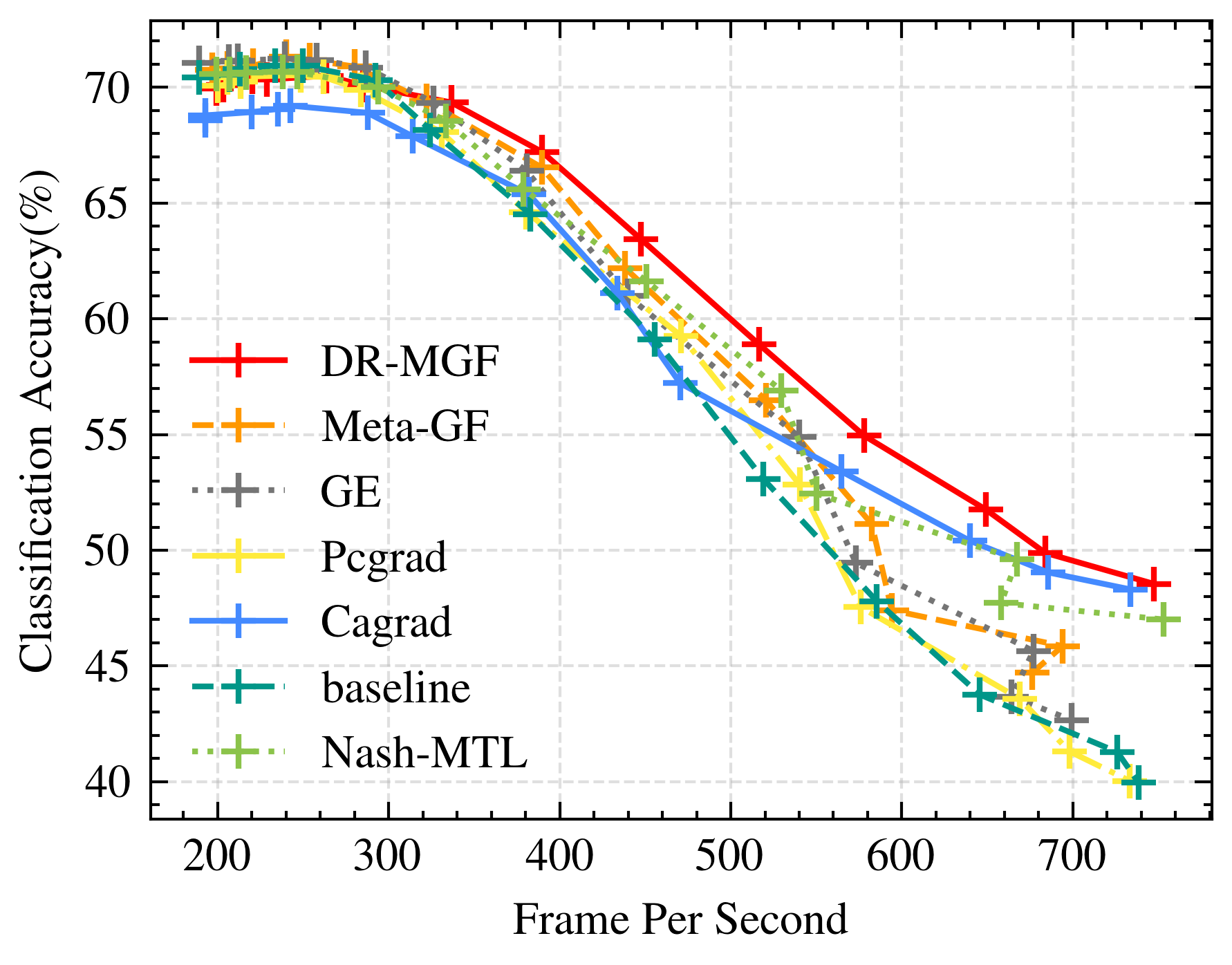}}
\subfigure[MSDnet]{
\includegraphics[height=40mm,trim=0 0 0 0,clip]{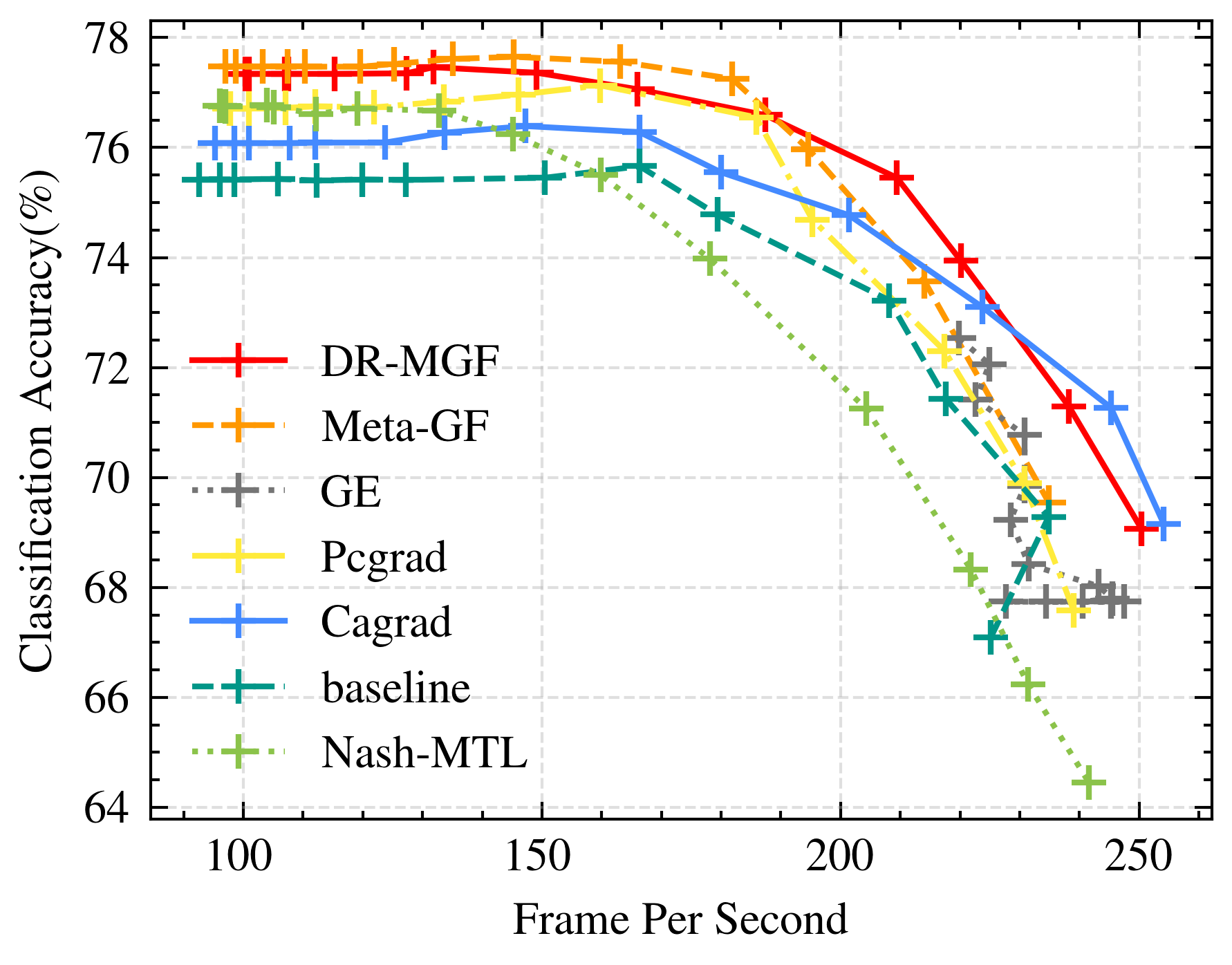}}
\caption{Performance comparison in depth-adaptive inference mode\cite{huang2017multi} (on CIFAR100). Classification accuracy is a function of average computational budget per image in this mode. The horizontal axis displays the different average inference speed, i.e., frames per second.}
\label{budget}
\end{figure*}

Different from gradient de-conflict approaches, recent works\cite{phuong2019distillation,wang2021harmonized,liu2020metadistiller} have developed knowledge distillation techniques to mitigate inter-task interference in multi-exit networks. We compare our approach with the distillation-based training method proposed in \cite{phuong2019distillation} as shown in Table \ref{KD}. The comparison results demonstrate that DR-MGF is complementary to the distillation-based methods, and can improve their performance. 
\begin{table*}[ht]
    \centering
     \caption{Multi-task learning results on NYUv2 dataset. \#P denotes the relative model size compared to the vanilla SegNet\cite{badrinarayanan2017segnet}. The best average result among all multi-task methods is marked in bold. DR-MGF outperforms baseline methods on three tasks.}
    \resizebox{0.75\textwidth}{!}{
    \begin{tabular}{ccrrrrrrrrrc}
    \toprule
     & &  \multicolumn{2}{c}{Segmentation} & \multicolumn{2}{c}{Depth} & \multicolumn{5}{c}{Surface Normal} & \\
    \cmidrule(lr){3-4}\cmidrule(lr){5-6}\cmidrule(lr){7-11}
     \#P. & Method &  \multicolumn{2}{c}{(Higher Better)} &  \multicolumn{2}{c}{(Lower Better)}  & \multicolumn{2}{c}{\makecell[c]{Angle Distance\\(Lower Better)}} & \multicolumn{3}{c}{\makecell[c]{Within $t^\circ$\\(Higher Better)}} & $\Delta m\% \downarrow$\\
     & & mIoU  & Pix Acc  & Abs Err & Rel Err & Mean & Median & 11.25 & 22.5 & 30 & \\ \midrule
     3 & Independent & 38.30 & 63.76 & 0.6754 & 0.2780 & 25.01 & 19.21 & 30.14 & 57.20 & 69.15 & \\ \midrule
     $\approx$3 & Cross-Stitch~ & 37.42 & 63.51 & 0.5487 & 0.2188 & 28.85 & 24.52 &  22.75 &  46.58 &  59.56 & 6.96\\
     $1.77$ & MTAN~\cite{liu2019end} & 39.29 & 65.33 & 0.5493 & 0.2263 &  28.15 &  23.96 &  22.09 &  47.50 &  61.08 & 5.59\\
     $1.77$ & MGDA~\cite{desideri2012multiple} &  30.47 &  59.90 &  0.6070 &  0.2555 & 24.88 & 19.45 & 29.18 & 56.88 & 69.36 & 1.38\\
     $1.77$ & Pcgrad~\cite{yu2020gradient} & 38.06 & 64.64 & 0.5550 & 0.2325 &  27.41 &  22.80 &  23.86 &  49.83 &  63.14 & 3.97\\
     $1.77$ & GradDrop~\cite{chen2020just} & 39.39 & 65.12 & 0.5455 & 0.2279 &  27.48 &  22.96 &  23.38 &  49.44 &  62.87 & 3.58\\
     $1.77$ & Cagrad\cite{liu2021conflict} & 39.79 & 65.49 & 0.5486 & 0.2250 & 26.31 & 21.58 & 25.61 & 52.36 & 65.58 & 0.20\\ 
     $1.77$ & Rotograd\cite{javaloy2021rotograd}& 39.32 & 66.07 & 0.5300 & 0.2100 & 26.01 & 20.80 & 27.18 & 54.02 & 66.53 & -2.31\\
     $1.77$ & Nash-MTL\cite{navon2022multi} & 40.13 &65.93 &0.5261&0.2171&25.26&20.08&28.40&55.47&68.15&-4.04 \\
     $1.77$ & Meta-GF\cite{sun2022meta} & 40.10 & 66.43 & 0.5176 & 0.2093 & 26.69 & 22.04 & 25.29 & 51.63 & 64.94 & -0.39\\ 
     $1.77$ & FAMO\cite{liu2023famo}  &38.88 &64.90& 0.5474& 0.2194& 25.06& 19.57& 29.21& 56.61& 68.98& -4.10\\
     $1.77$ & GO4Align\cite{shen2024go4align}& 40.42& 65.37 &0.5492 &0.2167 &24.76 &18.94& \textbf{30.54}& \textbf{57.87}&69.84 & -6.08\\
     \rowcolor{gray!40}
      $1.77$ &DR-MGF & \textbf{43.09} & \textbf{67.95} & \textbf{0.5073} & \textbf{0.2030} & \textbf{24.49} &\textbf{ 19.20} &30.29& 57.63 & \textbf{70.04} & \textbf{-8.39}\\ 
     \bottomrule 
    \end{tabular}
    }
    \label{tabnyuv2}
\end{table*}


\subsubsection{Performance comparison in depth-adaptive inference mode}
The multi-exit networks are designed to perform depth-adaptive inference, i.e., adjusting the inference depth according to the input complexity as shown in Fig. \ref{inferenceadaptive}. We adopt the budgeted batch prediction method\cite{huang2017multi} to compare the performance of different methods in adaptive inference mode. In this mode, the computational budget is given in advance and the multi-exit network is expected to allocate resources according to the complexity of inputs. For example, "easy" inputs are usually predicted by the shallow exits to save the computation resources. We refer the readers to the work proposed in \cite{huang2017multi} for more details. As shown in Fig. \ref{budget}, DR-MGF achieves superior performance on three types of models, demonstrating that DR-MGF can effectively strengthen the depth-adaptive inference performance of multi-exit networks.
\subsection{Performance on multi-task neural networks}
\begin{figure}[h]
\centering
    \includegraphics[width=0.45\textwidth]{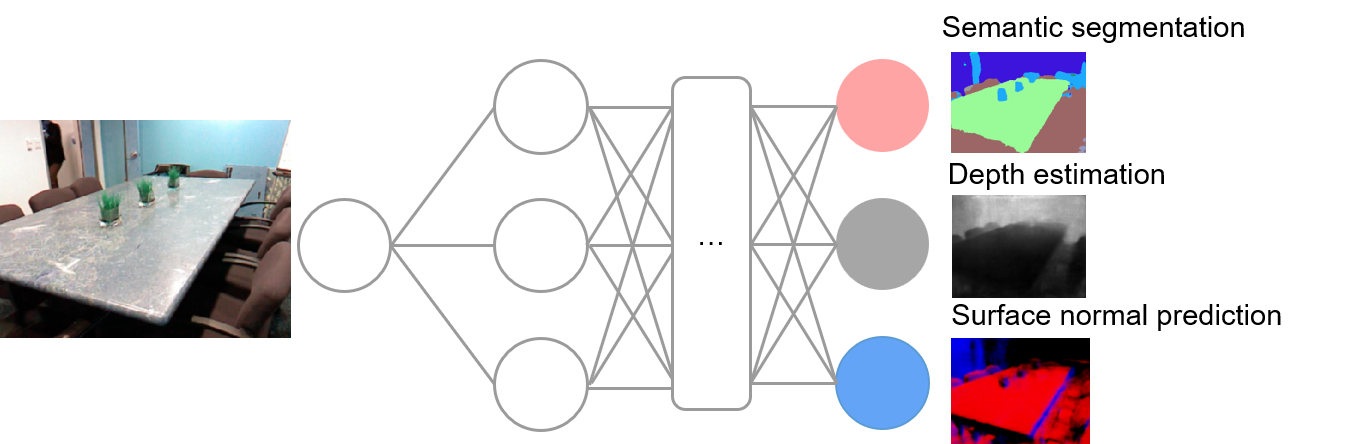} 
\caption{Schematic diagram of the prediction results of a multi-task network (NYUv2 dataset).}
\label{multi-taskprediction}
\end{figure}

Besides multi-exit neural networks, we also apply the proposed DR-MGF to the multi-task learning problem. We train the multi-task networks on NYUv2 dataset\cite{couprie2013indoor}. This dataset contains ground truth about the semantic segmentation task with 13 classes, depth estimation task and the normal surface prediction task. It consists of 1449 RGB-D images. Fig. \ref{multi-taskprediction} illustrates an example of the multi-task prediction results.

\begin{figure*}[t]
\centering
\subfigure[Vgg-SDN]{
\includegraphics[height=32mm,trim=0 0 0 0,clip]{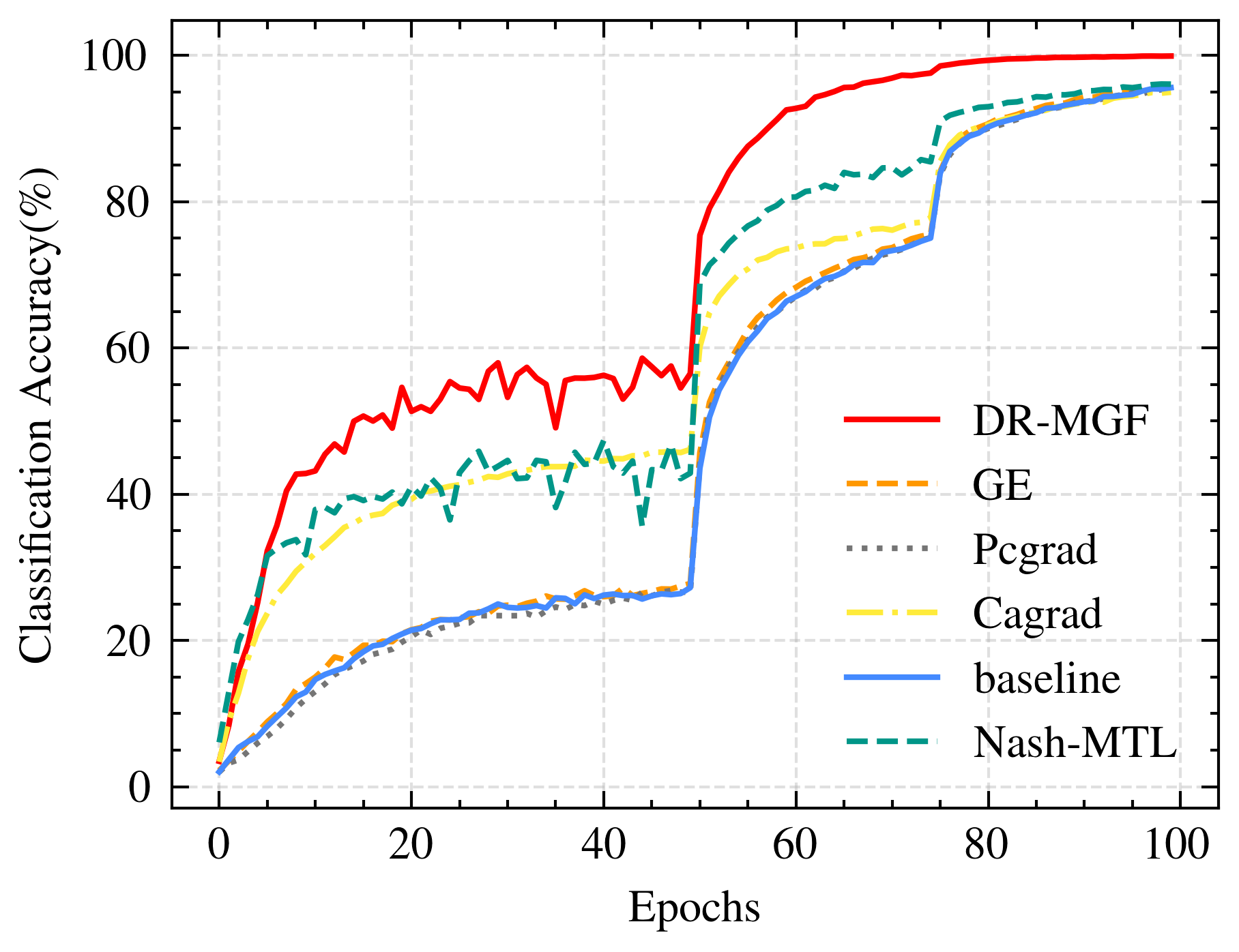}}
\subfigure[Vgg-SDN]{
\includegraphics[height=32mm,trim=0 0 0 0,clip]{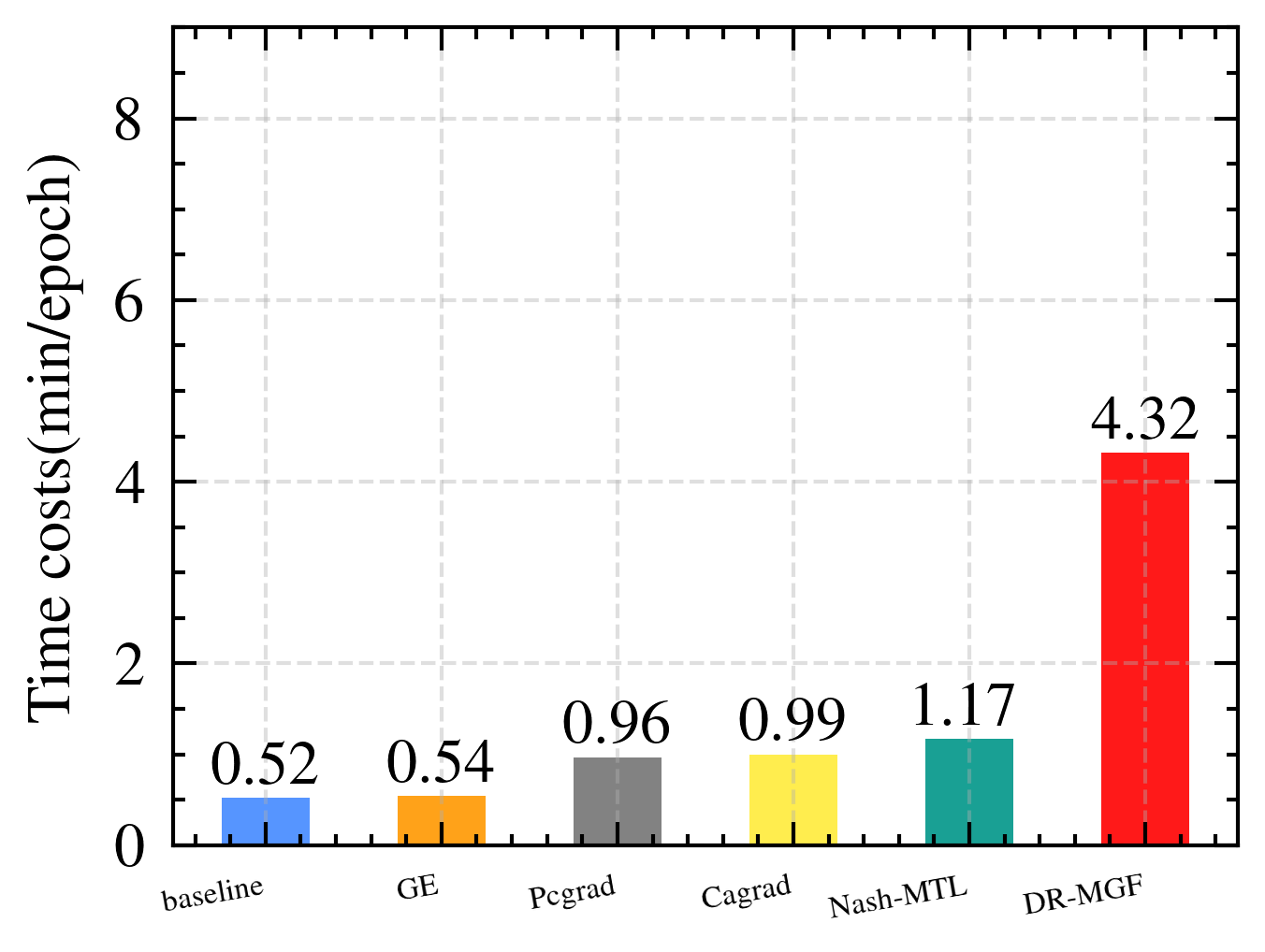}}
\subfigure[MSDnet]{
\includegraphics[height=32mm,trim=0 0 0 0,clip]{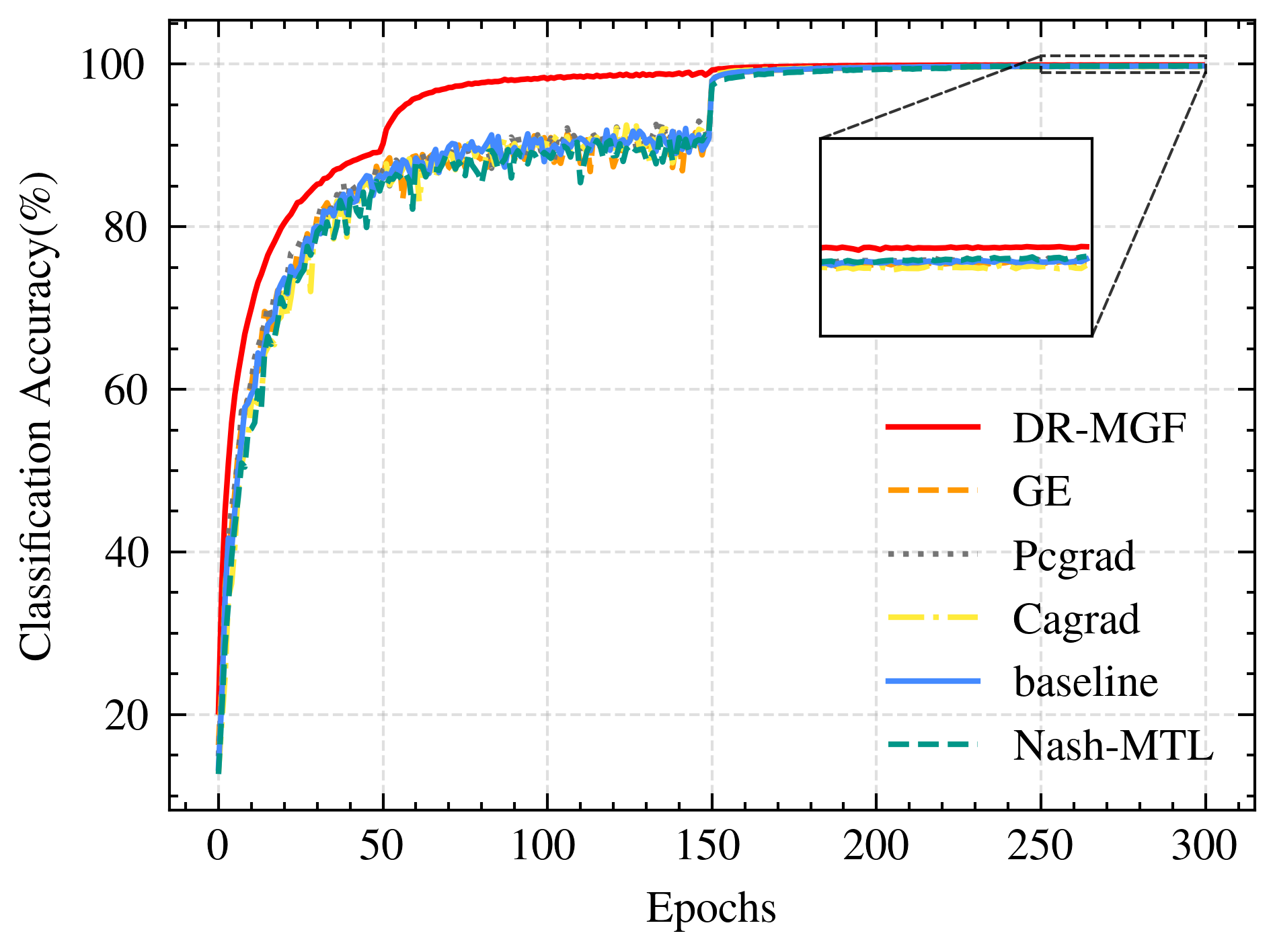}}
\subfigure[MSDnet]{
\includegraphics[height=32mm,trim=0 0 0 0,clip]{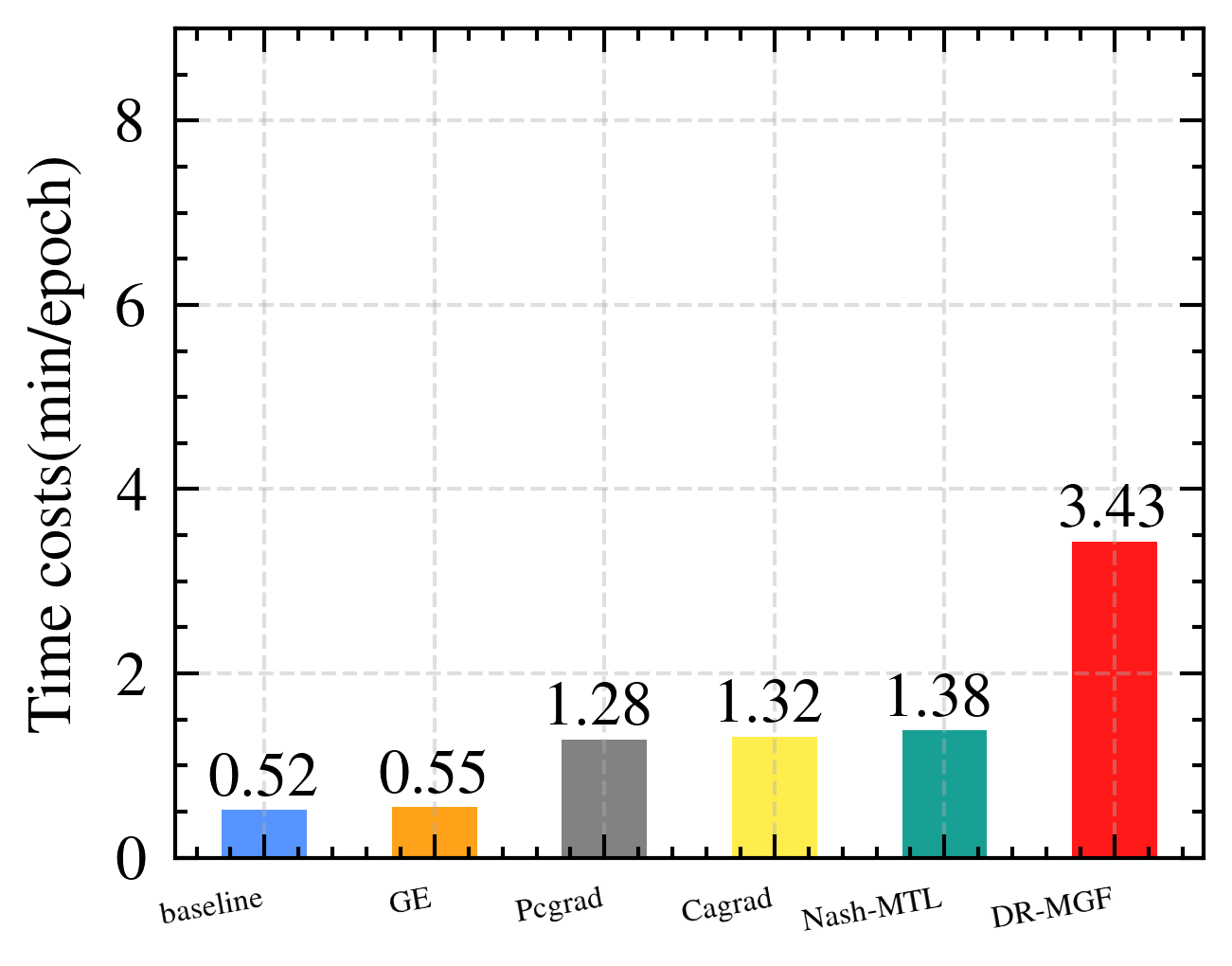}}
\caption{Comparison of different methods when applied to Vgg-SDN\cite{kaya2019shallow} and MSDnet\cite{huang2017multi} on CIFAR100. Subfigures (a) and (c) illustrate the average accuracy for each training epoch, while subfigures (b) and (d) demonstrate time cost per training epoch.}
\label{ablationconvergence}
\end{figure*}
 
\subsubsection{Implementation Details} 
We use Segnet-MTAN\cite{liu2019end} as the multi-task networks, which applies an attention mechanism on top of the SegNet\cite{badrinarayanan2017segnet} architecture. Following \cite{chen2018gradnorm}, we set the input image size to $288 \times 384$ and apply the same data augmentation policy. The baseline uses Adam optimizer with an initial learning rate of $1e-4$. The batch size is set to 2, and the maximum number of epochs is 200. We decay the learning rate by a factor of 0.5 at the 100th epoch. The test performance is the averaged results of the last 10 epochs. We compare the proposed DR-MGF with the state-of-the-art approaches including Pcgrad\cite{yu2020gradient}, Cagrad\cite{liu2021conflict} and GO4Align\cite{shen2024go4align}. The $\Delta m$ metric\cite{liu2021conflict} is employed to evaluate the overall performances of algorithm, which is defined as follows:
\begin{equation}
\label{deltam}
\centering
\Delta m=\dfrac{1}{K}\sum_{i=1}^{K} -(M_{m,i}-M_{0,i})/M_{0,i},
\end{equation}
where $i$ denotes the task index, $M_{0,i}$ denotes the single-task baseline of the $i$-th task, and $\Delta m$ represents the average performance degradation per-task using algorithm $m$ (smaller values indicate better performance).

\subsubsection{Comparison of performance in multi-task learning}
We evaluate the performance of each algorithm on three tasks: semantic segmentation, depth estimation and surface normal prediction. We repeat each method with 3 random seeds and the results are reported in Table \ref{tabnyuv2}. Compared with the single-task baseline, depth estimation benefits significantly from multi-task training process, while surface normal prediction suffers from inter-task interference. DR-MGF is designed to alleviate the gradient conflict problem of MONs, and enhance the performance of all tasks simultaneously. As demonstrated in Table \ref{tabnyuv2}, the interference of other tasks on surface normal prediction is notably reduced by DR-MGF, and DR-MGF achieves the best performance on the other two tasks at the same time. The results show that DR-MGF achieves the best overall performance with the lowest $\Delta m \%= -8.39$.

\subsection{Analysis about DR-MGF}
In this section, we first analyze the convergence results of MONs when using the proposed DR-MGF. Then we investigate whether the learned task-specific importance variables can reflect the importance of the shared filters for the corresponding tasks. 
\subsubsection{Comparison of the convergence results}
As shown in Fig. \ref{ablationconvergence}(a) and Fig. \ref{ablationconvergence}(c), DR-MGF improves the convergence performance of Vgg-SDN and MSDnet at each training epoch. Although the running time of DR-MGF is longer than other methods due to the disentanglement policy (Fig. \ref{ablationconvergence}(b) and Fig. \ref{ablationconvergence}(d)), DR-MGF can better alleviate inter-task interference, which is the fundamental motivation of this work. It enables various MONs converge to a better solution as illustrated in Tables \ref{cifartable_vgg}-\ref{tabnyuv2}. We have proposed two promising solutions in Sec. \ref{chapterconclusion} to reduce the computational overhead of DR-MGF.  \par
\begin{figure}[!htb]
\centering
\subfigure[]{\includegraphics[height=32mm,trim=0 0 0 0,clip]{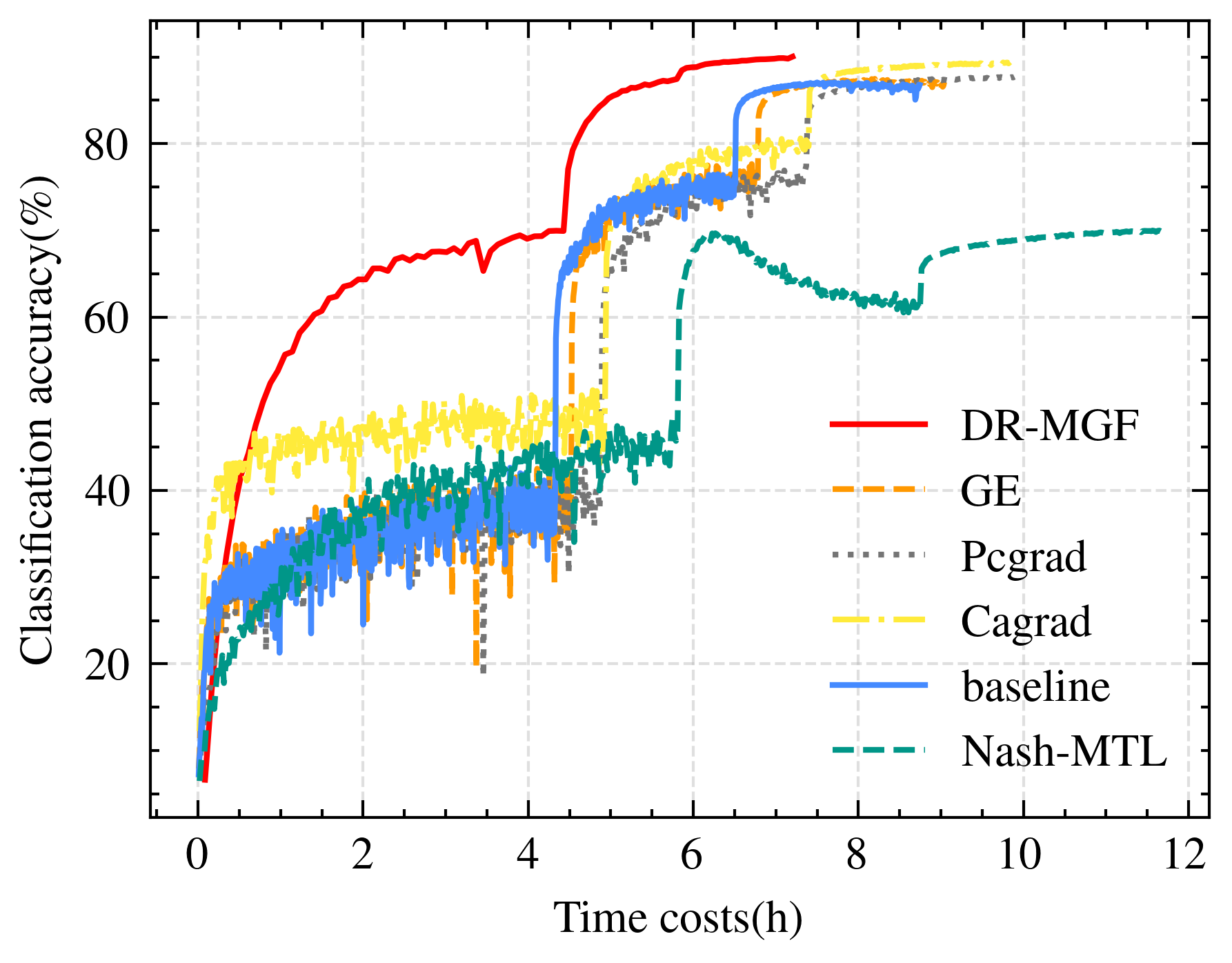}}
\subfigure[]{\includegraphics[height=32mm,trim=0 0 0 0,clip]{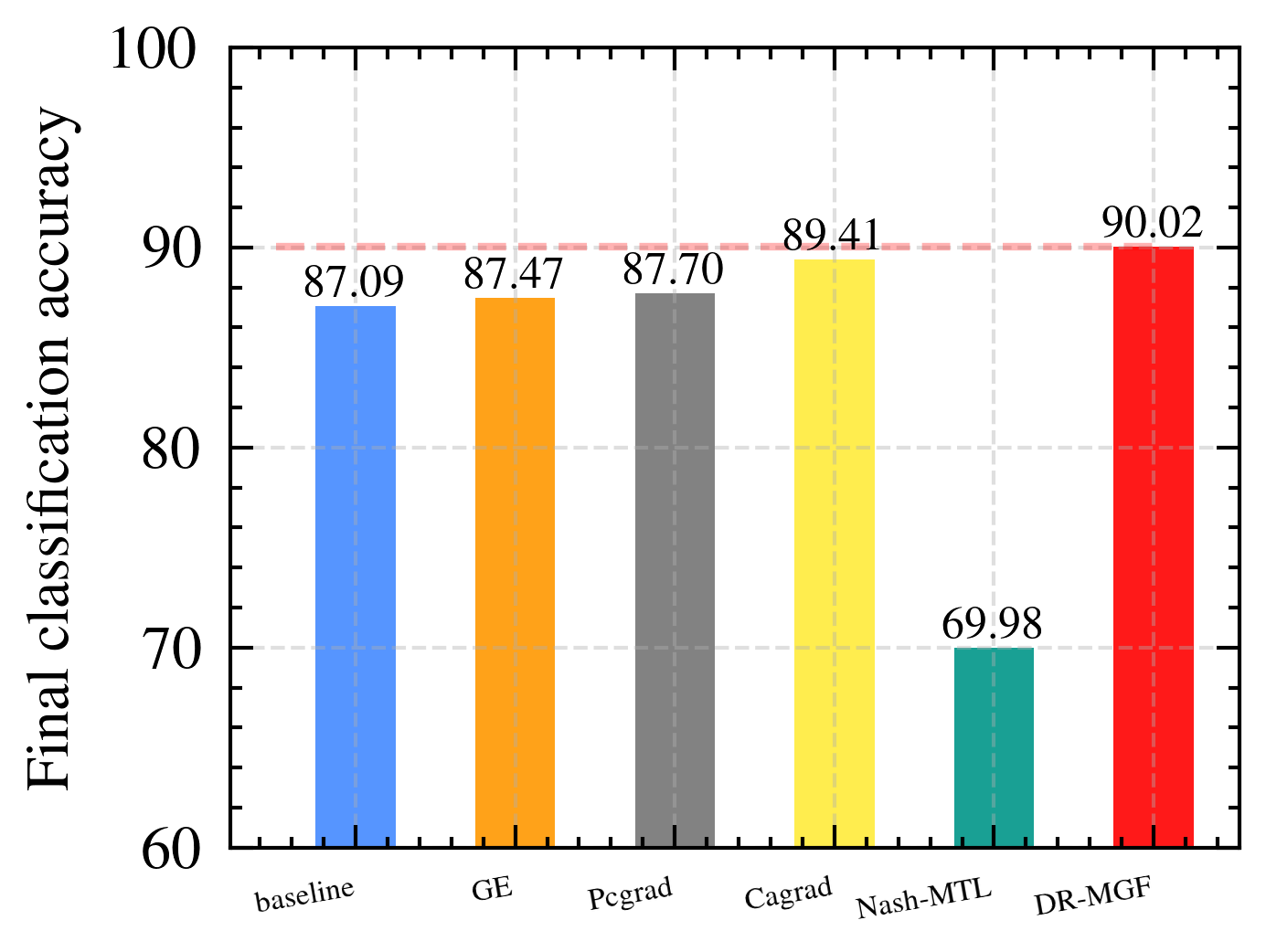}}
\caption{Comparison of final convergence results of Vgg-SDN on training sets of CIFAR100, ensuring the training time of existing methods is equal to or longer than that of the proposed DR-MGF.}
\label{equalcomparison}
\end{figure}

To let existing methods fully train the models, we further conduct experiments to compare each method by ensuring the training time of existing methods is equal to or longer than DR-MGF, i.e., increasing those methods' training epochs. Fig. \ref{equalcomparison} demonstrates that DR-MGF achieves better convergence, and still surpasses existing approaches. The detailed results are reported in Table \ref{timecostsablation}.\par
\begin{table}[h]
\centering
\caption{Comparison of training time and average accuracy on CIFAR100 (Vgg-SDN).}
\resizebox{0.5\textwidth}{!}{
\begin{tabular}{cc|cccccc}
\toprule
&Epochs&Vgg-SDN &GE\cite{li2019improved} &Pcgrad\cite{yu2020gradient}& Cagrad\cite{liu2021conflict}&Nash-MTL\cite{navon2022multi}&DR-MGF \\ \midrule
Running time (h)&\multirow{2}{*}{100}&0.8&0.9&1.6&1.65 &1.95&7.2  \\
Average accuracy&&66.34&66.19 &66.61&67.60 &64.29&\textbf{69.25} \\ \midrule
Running time (h)&\multirow{2}{*}{700}&8.74&9.03 &9.88&9.83 &11.66&- \\
Average accuracy&&66.50&66.76 &67.12&68.93 &54.28&-  \\
\bottomrule 
\end{tabular}
}
\label{timecostsablation}
\end{table}

\begin{figure}[htb]
\centering
\subfigure[]{
\includegraphics[height=33mm,trim=0 0 0 0,clip]{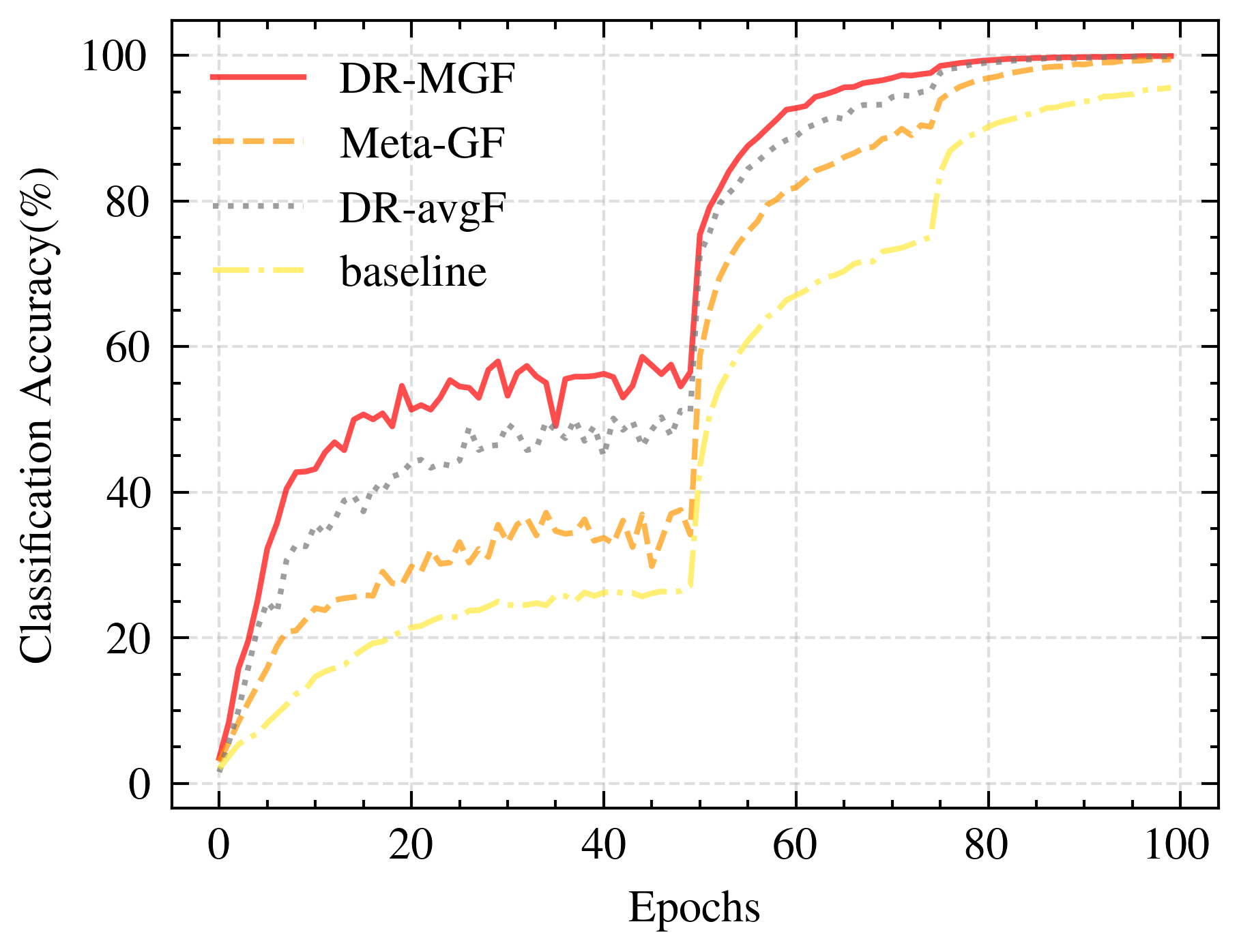}}
\subfigure[]{
\includegraphics[height=33mm,trim=0 0 0 0,clip]{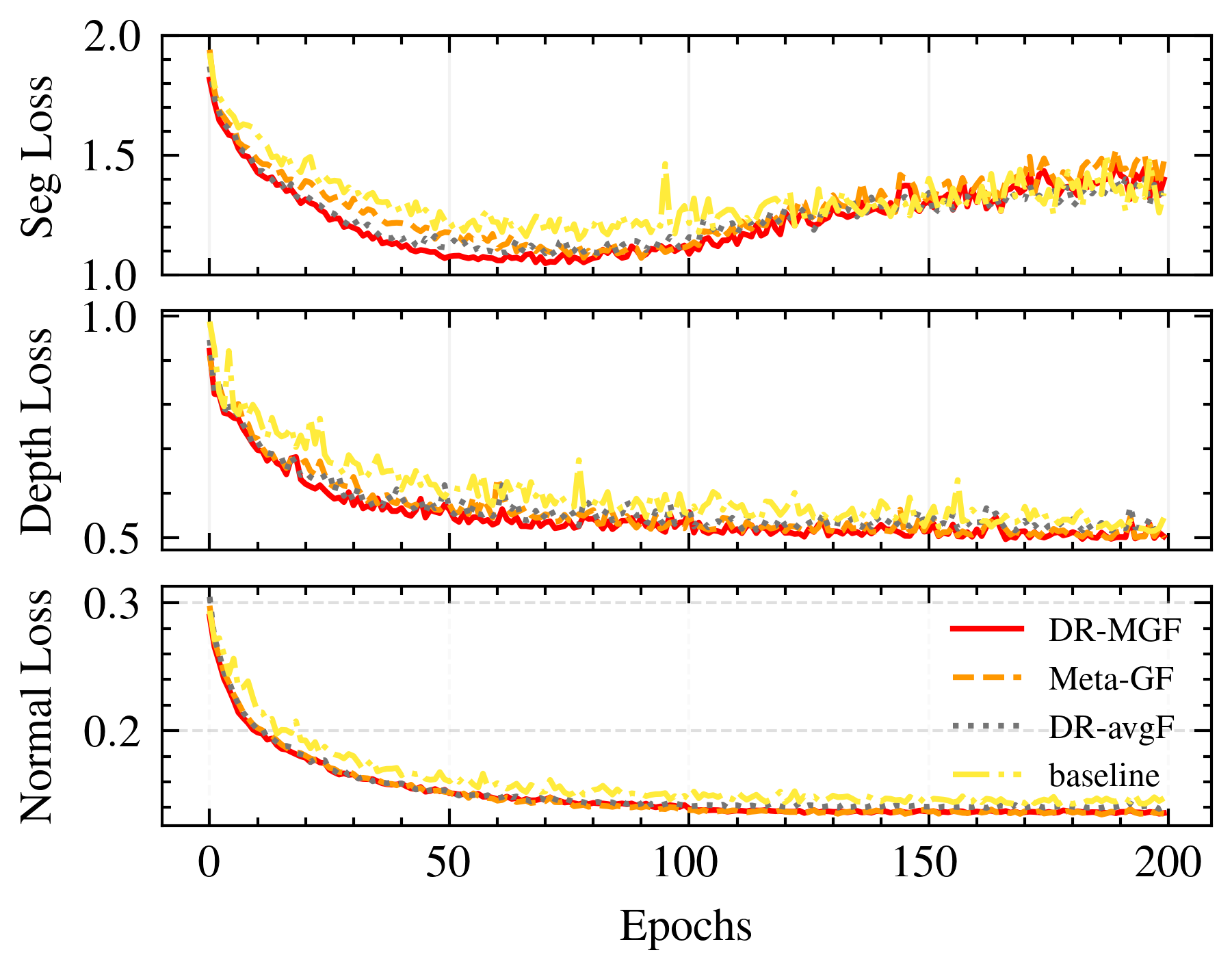}}
\caption{Comparative analysis of network convergence under different learning methods. (a) Classification accuracy curve of Vgg-SDN on CIFAR100; (b) Multi-task loss curves for segmentation, depth estimation and surface normal prediction of SegNet-MTAN on NYUv2.}
\label{ablationconvergence3}
\end{figure}

We further compare the convergence performance of the models by three ablation studies: 1) Meta-GF: using the meta-weighted gradient fusion policy without network disentanglement in the forward propagation; 2) DR-avgF: using vanilla average gradient fusion without meta-weighted gradient fusion policy, but optimizing the task-specific importance variables in the forward propagation; 3) DR-MGF: combining meta-weighted gradient fusion policy with network disentanglement. The baseline results are obtained using SGD. The results are shown in Fig. \ref{ablationconvergence3}(a)-(b), where DR-MGF achieves the best convergence performance.\par

\begin{figure*}[t]
  \centering 
\subfigure[]{\includegraphics[width=0.35\textwidth]{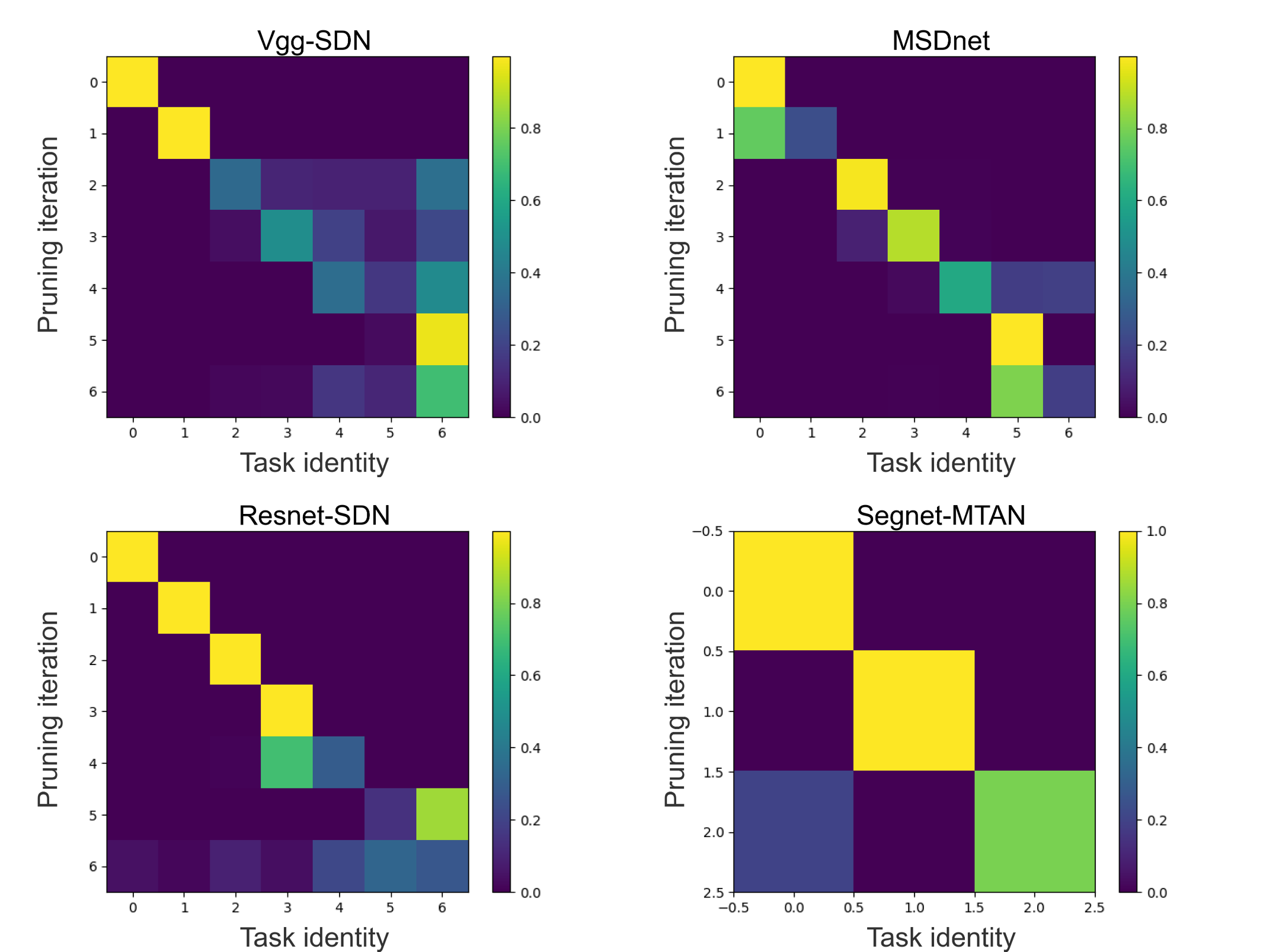}} 
\subfigure[]{\includegraphics[width=0.62\textwidth]{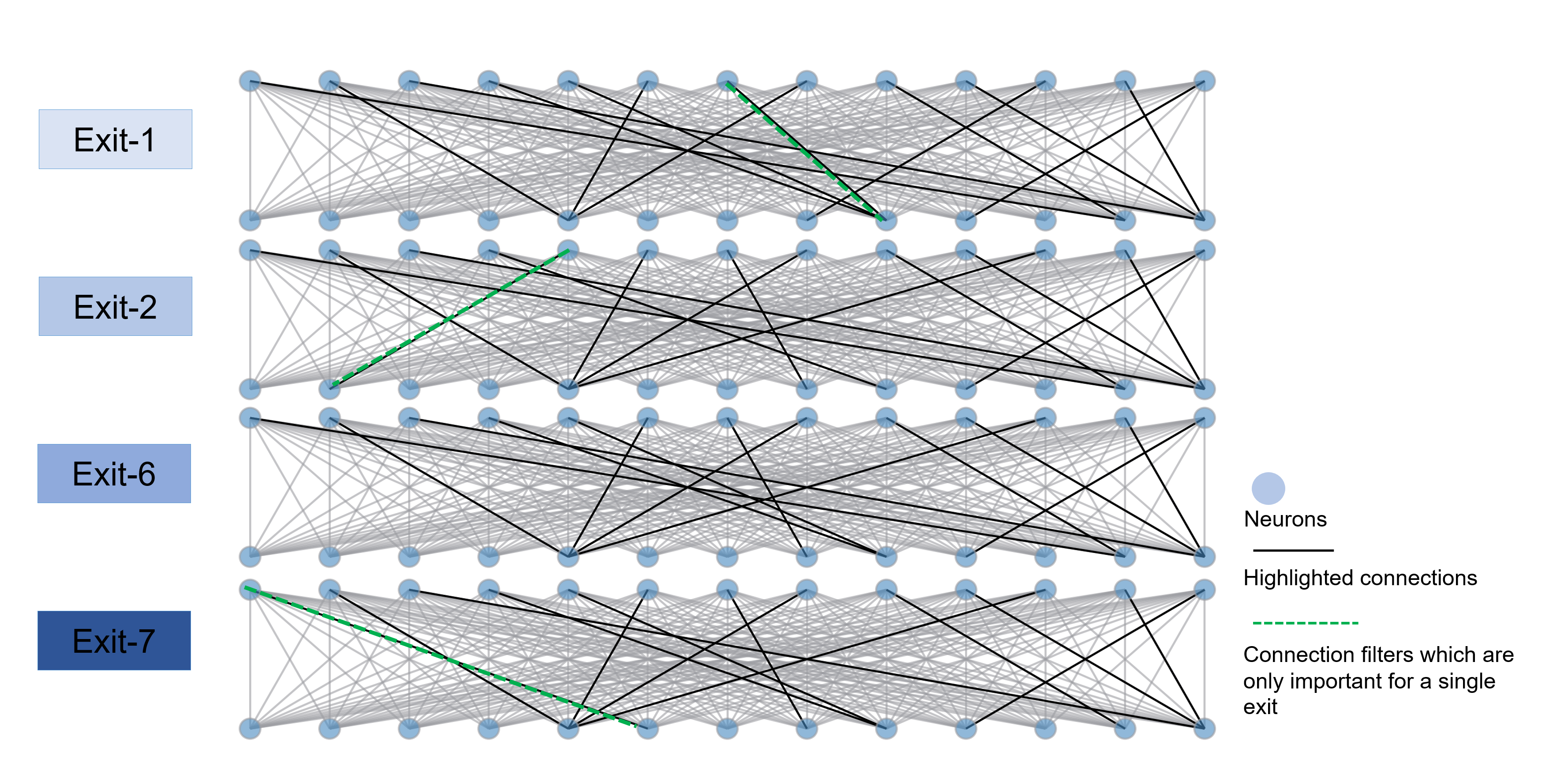}}
\caption{Analysis of the learned task-specific importance variables. (a) The accuracy degradations when pruning the important shared filters of different tasks; (b) Visualization of the forward task-preferred structure (CIFAR100,Vgg-SDN). For brevity,  we highlight top $20\%$ connection filters with higher task-specific importance variables between the 1st and 2nd layers of Vgg-SDN.}
\label{pruning}
\end{figure*}

\subsubsection{Analysis of the learned task-specific importance variables}
\label{learnedfusionweight}
We further make analysis of the learned task-specific importance variables $\nu$ on CIFAR100 and NYUv2. We preliminarily define the $i$-th parameter $w$ with the largest $\nu_{k,i}$ as the task-specific important filter for the $k$-th task, where $\{\nu_{k,i}|\nu_{k,i}>\nu_{k^{'},i},k\in[1,K],k\neq k^{'}\}$. For multi-exit networks, we iteratively prune the important parameters of each task from the $1$st exit to the $7$th exit. For multi-task networks, we iteratively prune the important parameters of each task from the $1$st task to the $3$rd task on NYUv2 dataset.\par
As shown in Fig. \ref{pruning}(a), the relative accuracy degradation is plotted along the horizontal axis, while the vertical axis represents the pruning iteration. It can be seen that when we prune the important parameters of one task, it primarily reduces the accuracy of that task, indicating that the learned task-specific importance variables effectively capture the importance of shared parameters to each task. \par
We further visualize the sub-structures shaped by task-specific importance variables. For clarity, we highlight the top $20\%$ connection filters with higher task-specific importance variables. Fig. \ref{pruning}(b) demonstrates that the 1st, 2nd and 7th exits of Vgg-SDN own their own preferred inference routes. Additionally, the task-preferred inference route of the 6th exit overlaps significantly with the 7th exits. This aligns with the result demonstrated in Fig. \ref{pruning}(a), where pruning the important filters of the 7th exit causes significant performance degradation of the 6th exit.

\section{Conclusion}
\label{chapterconclusion}
In this work, we propose DR-MGF to alleviate the gradient conflict problems when training MONs. Unlike existing approaches, DR-MGF alleviates gradient conflicts among tasks from the perspective of network disentanglement. By learning the task-specific importance variables, each task automatically identifies its preferred filters during the disentanglement stage. Then in the fusion stage, DR-MGF employs a meta-weighted gradient fusion policy to integrate task gradients based on the learned task-specific importance variables. Through integrating disentanglement and fusion stages, DR-MGF enables tasks to dominate the optimization of their preferred  filters, and finally shape task-preferred inference sub-structures at the end of training. We provide a detailed analysis of the proposed approach, and conduct extensive experiments on CIFAR, ImageNet and NYUv2. The experimental results demonstrate the superiority of our approach. Additionally, the disentanglement-and-fusion policy might be useful for alleviating the catastrophic forgetting problem in task-incremental continual learning, making it worthwhile to extend DR-MGF to the continual learning of deep neural networks.\par
Although DR-MGF achieves the best performance on various MONs, the running time of DR-MGF can be further reduced in the future. According to our preliminary study, there are two promising solutions:
\begin{itemize}
\item  Adaptively merging tasks with slight conflicts into the same task group will reduce the number of disentanglement stages, and improve the training efficiency of DR-MGF while maintaining a good overall performance of MONs.
\item  Employing DR-MGF only when inter-task interference exceeds a threshold would significantly reduce training computation and storage consumption.
\end{itemize}

\renewcommand{\theequation}{\thesection.\arabic{equation}}
\setcounter{equation}{0}
\appendices
\section{}
\label{appendix_toy}
To assess the capability of DR-MGF and existing methods in alleviating gradient conflicts, we design a two-task neural network $M$ in the toy experiment as illustrated in Fig. \ref{overview} of main manuscript. The two task models $\{\theta_{1},\theta_{2}\}$ are composed of six learnable weights: $\{w_{1},w_{2},a_{1},a_{2},b_{1},b_{2}\}$, namely:  
\begin{equation}  
\begin{aligned}  
\theta_{1} &= \text{sign}(a_{1}) \odot \left|a_{1}\right|w_{1} + \text{sign}(b_{1}) \odot \left|b_{1}\right|w_{2}, \\  
\theta_{2} &= \text{sign}(a_{2}) \odot \left|a_{2}\right|w_{1} + \text{sign}(b_{2}) \odot \left|b_{2}\right|w_{2},  
\end{aligned}  
\end{equation}  
where $\{w_{1},w_{2}\}$ are the weights of shared filters. As described in Eq. (\ref{losssurfacetoy}), $\{f_{1}(\theta), f_{2}(\theta)\}$ represent the objective functions of both tasks:  
\begin{equation}  
\centering  
\begin{aligned}  
f_{1}(\theta) &= \text{max}\left\{ (\theta+[c_{1},c_{1}])((\theta+[c_{1},c_{1}]))^{T}, c_{1}\sqrt{2} \right\} \\
&+ c_{2}(\theta+[c_{1},c_{1}])(\theta+[c_{1},c_{1}])^{T}, \\  
f_{2}(\theta) &= \text{max}\left\{ (\theta-[c_{1},c_{1}])((\theta-[c_{1},c_{1}]))^{T}, c_{1}\sqrt{2} \right\} \\
&+ c_{2}(\theta-[c_{1},c_{1}])(\theta-[c_{1},c_{1}])^{T},  
\end{aligned}  
\label{losssurfacetoy}  
\end{equation}
where constants $c_{1}$ and $c_{2}$ are empirically set to 8 and 0.4, respectively. Both variables can be set to other values.

\section{} 
\label{appendix_conflict}

The loss degradation of $f_{1}$ can be approximated by a first-order Taylor expansion\cite{yu2020gradient} when the learning rate $\epsilon$ is small:
\begin{align}
\Delta f_{1}^{g_{1}}&=f_{1}(w_{0}-\epsilon g_{1})-f_{1}(w_{0})\\
&\approx f_{1}(w_{0})-\nabla_{w}f_{1}(w_{0})(\epsilon g_{1})+o(\epsilon^{2})-f_{1}(w_{0}) \\
&=-\epsilon g_{1}^{T}g_{1}+o(\epsilon^{2}),
\end{align}
when applying the joint gradient $g_{1}+g_{2}$ to update $w_{0}$, the loss degradation can be calculated by:
\begin{align}
\Delta f_{1}^{g_{1}+g_{2}}&=f_{1}(w_{0}-\epsilon (g_{1}+g_{2}))-f_{1}(w_{0})\\
&\approx -\epsilon (g_{1}^{T}g_{1}+g_{1}^{T}g_{2})+o(\epsilon^{2}), \\
\end{align}
therefore, when $g_{1}$ conflicts with $g_{2}$, i.e., $g_{1}\cdot g_{2}<0$:
\begin{align}
\Delta f_{1}^{g_{1}+g_{2}}< \Delta f_{1}^{g_{1}} \qquad g_{1}^{T}g_{2}< 0.
\end{align}

\section{}
\label{appendix_relation}

The relative convergence gain can be expanded as:
\begin{align}
G&=\dfrac{\Delta f_{1}^{g_{1}+g_{2}}-\Delta f_{1}^{g_{1}}}{\Delta f_{1}^{g_{1}}}\\
&\approx \dfrac{-\epsilon \left(g_{1}^{2}+g_{1}g_{2}\right)+o(\epsilon^{2})-(-\epsilon \left(g_{1}^{2}\right)+o(\epsilon^{2}))}{-\epsilon \left(g_{1}^{2}\right)+o(\epsilon^{2})}\\
&=\dfrac{-\epsilon \left(g_{1}g_{2}\right)+o(\epsilon^{2})}{-\epsilon \left(g_{1}^{2}\right)+o(\epsilon^{2})},
\end{align}
therefore, we calculate the inter-task gradient conflict value $C$ by:
\begin{equation}
C=\sum_{i=1}^{N}max(0,\dfrac{-g_{1,i}^{T}g_{2,i}}{\|g_{1}\|^{2}}).
\end{equation}

\section*{Acknowledgement}
This work is supported by the National Natural Science Foundation of China U24A20279, 61973311 and the STI 2030-Major Projects (2022ZD0208903).

\bibliographystyle{IEEEtran}
\bibliography{IEEEfull,reference}

\end{document}